\documentclass[arxiv]{meowreport}

\usepackage{algorithm}
\usepackage{algpseudocode}
\usepackage{graphicx}
\usepackage{makecell}
\usepackage{booktabs}
\usepackage[accsupp]{axessibility}
\usepackage{xcolor}
\usepackage{algorithm}
\usepackage{algpseudocode}
\usepackage{amsmath}
\usepackage{bbding}
\usepackage{float}
\usepackage{multirow}

\meowtitle{PIPBench: A Profile-Inclusive Framework for Personalized Image Generation Evaluation}
\meowauthors{Yuhang Wu\affmark{1} \quad
Shuxiang Zhang\affmark{2} \quad
WEE HIAN CHING\affmark{1} \quad
Chi Zhang\affmark{1,3} \quad
Miao Liu\affmark{1}\corrauth}
\meowaffiliation{\affmark{1}College of AI, Tsinghua University\\
\affmark{2}Sun Yat-sen University\\
\affmark{3}Shanghai Qi Zhi Institute\\
}
\meowdate{July 2, 2026}
\meowlinks{
  \meowlink{Project Page}{https://wuyuhang05.github.io/PIPBench/}
  \quad
  \meowlink{Code}{https://github.com/wuyuhang05/PIPBench}
  \quad
  \meowlink{Dataset}{https://huggingface.co/datasets/AirRain03/PIPBench}
}

\begin{document}

\makemeowtitle

\vspace{-2em}
\begin{meowabstract}
Recent text-to-image models such as DALL·E-3 excel at following diverse prompts yet remain blind to individual aesthetic preferences. We study personalized image generation, where models must align outputs with a user’s implicit visual preferences based on a few historically preferred images and a short prompt. To this end, we introduce PIPBench, the first profile-inclusive benchmark for evaluating personalized image generation. We further propose a novel data construction pipeline that leverages psychological and demographic profiling dimensions for both real-user data collection and scalable agent-based data generation. Using PIPBench, we conduct a thorough evaluation of representative line of methods. Our experiments reveal key limitations in existing methods, suggesting new challenges and opportunities for personalized text-to-image synthesis.
\end{meowabstract}

\vspace{-2em}
\begin{meowteaser}
  \includegraphics[width=\textwidth]{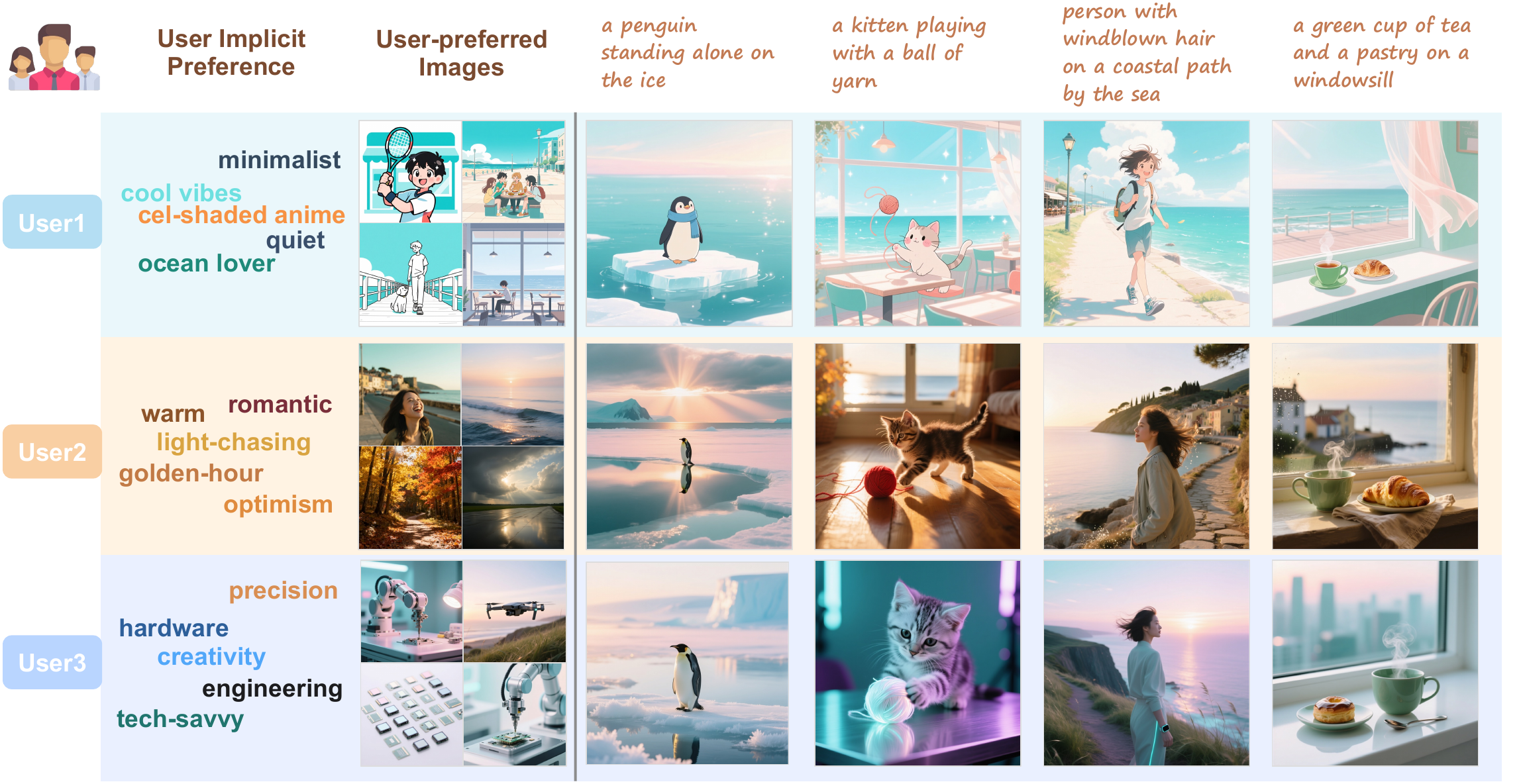}
  \caption{\textbf{Problem setup of personalized image generation and examples from our PIPBench}. Our benchmark incorporates both user profiles and preferred images, enabling a comprehensive analysis of implicit and explicit visual aesthetic preferences.}
  \label{fig:teaser}
\end{meowteaser}

\section{Introduction}
\label{sec:intro}



Recent text-to-image generation models, such as DALL·E 3~\cite{betker2023improving}, have attracted widespread attention. However, these systems~\cite{betker2023improving,wu2025qwen,dai2023emu} remain largely task-driven: they readily follow explicit user instructions, yet adopt a generic generative model without accounting for implicit individual preferences. 

Consider the examples in Fig.~\ref{fig:teaser}, a typical user prompt conveys only the core concept. Existing systems first expand such prompts using large language models (LLMs), and then generate images based on enriched descriptions. This two-stage procedure introduces an enormous search space due to the stochasticity of both language enrichment and image synthesis. However, each user may resonate with only a small portion of this exploration space, resulting in a high interaction barrier for non-expert users, since personal preferences are often implicit and difficult to express through plain language.

An alternative paradigm for personalized text-to-image generation is illustrated in Fig.~\ref{fig:teaser}. Specifically, the generation model leverages historical user preference data to produce results that are better aligned with the user’s aesthetic intent. This setting relates to prior work on style transfer and image editing, which align visual synthesis with visual or textual guidance. However, these methods do not consider user aesthetic preference modeling.

Notably, only a handful recent studies~\cite{xu2025personalized,salehi2024viper,kim2025draw} have examined image generation conditioned on user preference data, largely due to the absence of a systematic benchmark with high-quality user data. Existing datasets~\cite{chen2024tailored,chen2019pog,harper2015movielens} that include user interaction histories are largely domain-specific, focusing on applications such as sticker or poster preferences. In contrast, for general personalized image generation, individual implicit visual preferences are inherently difficult to articulate and quantify. For example, a user may consistently favor images depicting open natural landscapes over enclosed urban scenes, without being consciously aware of this tendency. Such inclinations are challenging to collect, as they are diverse and subjective, manifested through behavioral or profiling patterns~\cite{bourdieu2019distinction,mccrae2007aesthetic} rather than explicit statements. However, user profiling information is often overlooked in existing image generation works, leaving these implicit preference signals underrepresented.




To bridge this gap, we introduce \textbf{PIPBench}, the \textbf{P}rofile-\textbf{I}nclusive \textbf{P}ersonalized Image Generation \textbf{Bench}mark. Drawing on established psychological literature, we first formalize profile axes that implicitly shape visual preferences. These axes guide the design of a structured survey for real-user profile collection and enable the scalable construction of synthetic agents to diversify benchmark data distribution and facilitate large-scale training data generation. We then generate candidate image sets for preference selection by prompting Large Language Models (LLMs) to produce profile-conditioned captions, followed by image generation. We evaluate representative prior methods on PIPBench, explore training-based methods for preference conditioning, and validate the benefits of our profile-inclusive framework. Our contributions are summarized as follows:
\begin{meowcontributions}
    \item We introduce the first personalized image generation benchmark, featuring real user profiles paired with their corresponding preferred images.
    \item We propose a novel agent-based pipeline for diverse benchmark data construction and scalable training data generation.
    \item We conduct comprehensive analyses of prevailing personalized image generation methods, providing insights into their strengths and limitations.
\end{meowcontributions}

\section{Related Work}
\label{sec:rel}



\noindent\textbf{Personalized Modeling and Generation}.\
Our work is most closely related to existing studies on personalized generation. Prior research has formed two paradigms. The first is the test-time fine-tuning paradigm: optimizing model parameters specifically for each user or concept can achieve strong personalized effects, but it is accompanied by significant computational and storage overhead, making it difficult to scale in large-scale service scenarios \cite{ruiz2023dreambooth}. The other, more relevant paradigm leverages conditional projection to extract conditioning information from reference images using foundation models~\cite{Patashnik_2021_ICCV, salehi2024viper, kim2025draw, xu2025personalized, li2025instantpreferencealignmenttexttoimage, chen2024tailored, mo2025prefgen}. Salehi \etal~\cite{salehi2024viper} proposed to mine core features of user styles from reference images via contrastive learning, converts them into conditional vectors, and injects them into the cross-attention modules of generative models to achieve personalized binding of style and content; Li \etal~\cite{li2025instantpreferencealignmenttexttoimage} employs multimodal large language models to mine global preference signals from reference images, enabling personalized generation without additional training. Kim \etal~\cite{kim2025draw} incorporates user interaction history into condition modeling in the latent space through Transformer adapters, enabling adaptation to mainstream foundation models without additional fine-tuning. Chen \etal~\cite{chen2024tailored} leverages historical user prompts to personalized prompt rewriting via LLMs. Importantly, existing works have not addressed the implicit user preferences embedded in user profiles, primarily due to the limitations of existing benchmarks.


\noindent\textbf{Image Generation Benchmarks}.\
Existing evaluation systems for personalized generation are mainly constructed around fidelity, alignment, and semantic consistency, and have played an important role in object-level personalized assessment. DreamBench++~\cite{ICLR2025_71ad539a} introduces a personalized image-generation benchmark that uses a multimodal GPT to automatically align with human preferences, aiming to improve the agreement between automated evaluations and human subjective judgments. Meanwhile, some studies have begun to introduce subjective metrics such as user satisfaction to supplement objective evaluations~\cite{NEURIPS2022_ec795aea, mo2025learninguserpreferencesimage, xu2023imagereward, NEURIPS2023_73aacd8b}. Xu \etal\cite{xu2023imagereward} constructs the ImageRewardDB dataset, which includes a substantial number of expert-annotated image comparison pairs and is used to train human preference reward models for text-to-image generation. Kirstain \etal~\cite{NEURIPS2023_73aacd8b} constructs a large-scale open dataset named Pick-a-Pic, which collects real preference feedback from image generation enthusiasts via a dedicated web application and includes a vast number of image comparison pairs generated based on diverse text prompts. 
To specifically evaluate personalized generation capabilities, Chen \etal~\cite{chen2024tailored} propose a history-based benchmark that takes a pool of user historical prompts as input to assess how well models align with individual intentions. Salehi \etal~\cite{salehi2024viper} introduce the Viper data set, which uses user comments alongside reference images to systematically measure a model's ability to capture user-specific stylistic preferences. These previous benchmarks largely rely on synthetic data or isolated contextual cues—such as relying solely on a pool of prompts, scattered user comments, or specific reference images. Even the few datasets that incorporate real user feedback typically lack comprehensive demographic information and holistic user profiles, resulting in limited controllability over data quality and a failure to capture implicit, multi-dimensional user preferences. We provide a detailed comparison between our benchmark and existing ones in \cref{tab:compare_benchmark} of the Supplementary Material.



\noindent\textbf{Textual Inversion and Style Transfer}.\
Previous studies on text embedding adaptation and diffusion-based style transfer laid the foundation for personalized image generation~\cite{gal2022image, ruiz2023dreambooth, hierarchicaltextconditionalimagegeneration, NEURIPS2022_ec795aea, ArtBank, Chung_2024_CVPR}. Text embedding methods achieve the encoding of target styles or objects by learning compact word vectors from a small number of reference images to represent specific concepts. Ramesh \etal~\cite{hierarchicaltextconditionalimagegeneration} proposes a hierarchical text-conditional image generation method that captures the target style and semantics by leveraging CLIP's pre-trained representations to achieve image generation with high diversity and high photorealism
; diffusion-based style transfer accomplishes pixel-level style alignment through mechanisms such as cross-attention. Chung \etal~\cite{Chung_2024_CVPR} proposes a training-free diffusion-based style transfer method, achieving an accurate fusion of content and style by replacing self-attention key-value pairs. Zhang \etal~\cite{ArtBank} Combines pre-trained diffusion models with an implicit style prompt bank, realizing high-quality artistic style transfer while preserving content structure. 
However, these methods typically focus on a single concept or style, struggling to effectively integrate cross-image co-occurrence cues present in multiple reference images, thus having significant limitations in capturing implicit user preferences.


\section{Benchmark Construction Pipeline}
\label{sec:data}

\begin{figure*}
    \centering
    \includegraphics[width=1\linewidth, keepaspectratio]{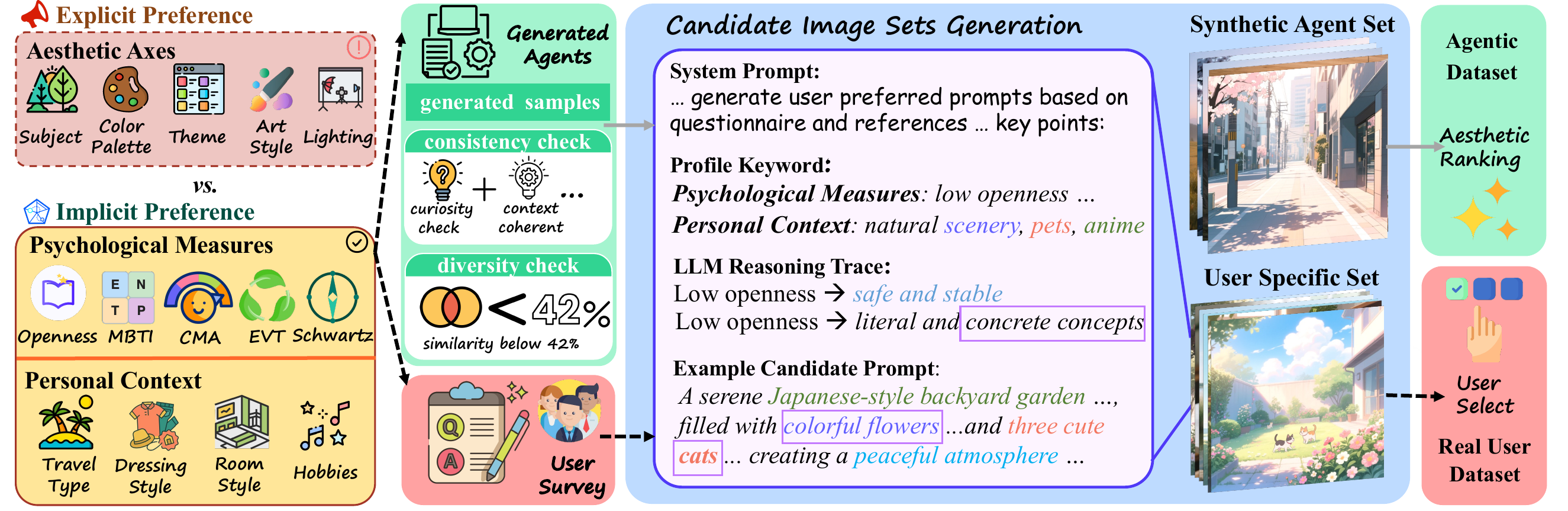}
    \caption{\textbf{Overview of the PIPBench data collection pipeline}.  We first define base user profiles using psychological measures and personal context. Real user profiles are collected through surveys, while synthetic agent profiles are generated by sampling from the profiling schema, followed by rigorous consistency and diversity checks. We then prompt Large Language Models (LLMs) to generate profile-aligned image prompts by reasoning over the relationship between latent user profiles and explicit aesthetic dimensions. Finally, given candidate image sets, preferred images selected by real users form a human-calibrated Real-User Dataset, while a synthetic Agentic Dataset is constructed automatically via aesthetic-score ranking within each set.} 
    \label{fig:pipeline}
\end{figure*}

We adopt a synthetic–real hybrid design, where high-cost real-user data collection provides reliable evaluation resources, while a scalable agent-based engine both enriches benchmark diversity beyond the collected data and enables large-scale training data generation. A visual illustration of our benchmark construction process is provided in \cref{fig:pipeline}. Specifically, we design profiling axes that encode implicit visual preferences. Guided by these axes, we collect real-user profile data and construct synthetic agents. We then generate profile-conditioned candidate image sets. The final benchmark includes real-user selections as well as a synthetic component constructed via automatic ranking.

\subsection{User Profile Definition}
\label{sec:prodef}

Unlike explicit aesthetic attributes, implicit visual preferences are inherently shaped by personal experience. Notably, such preferences are abstract and diverse, making them difficult to reliably articulate or systematically collect through direct surveys or discrete selection tasks. Nevertheless, established psychological studies suggest that such implicit inclinations are systematically correlated with individuals’ psychological traits~\cite{mccrae2007aesthetic} and contextual backgrounds~\cite{bourdieu2019distinction}.

\noindent\textbf{Psychological Measures}.\
User visual preferences arise from a combination of personality traits and individual dispositions that jointly shape aesthetic judgment. Based on a comprehensive review of the psychological literature, we consider the following psychological measures:

\begin{itemize}
\item \textbf{Big Five Model}~\cite{Big5,Openness}.\ Defines personality along five dimensions: \textit{Openness, Conscientiousness, Extraversion, Agreeableness,} and \textit{Neuroticism}. Among these, \textbf{Openness} is most relevant to visual preference~\cite{Openness}, reflecting imagination, aesthetic sensitivity, and receptivity to novelty. Individuals high in Openness favor abstract and unconventional imagery, whereas those low in Openness prefer realistic, structured compositions.

\item \textbf{Ten-Item Personality Inventory (TIPI)}~\cite{TIPI,TIPI}.  
A concise measure of the Big Five traits. We use it to estimate \textbf{Openness}, where higher scores indicate curiosity and preference for visual complexity and novelty.  

\item \textbf{Schwartz’s Basic Human Values}~\cite{Schwartz}.  
Defines ten motivational values shaping human goals and artistic taste. High \textit{Self-Direction} or \textit{Stimulation} aligns with innovative, exploratory aesthetics, while high \textit{Tradition} or \textit{Security} corresponds to classical and balanced compositions.  

\item \textbf{Ecological Valence Theory (EVT)}~\cite{EVT}.  
Explains color preference through affective associations with real-world experiences. Warm palettes evoke comfort and nostalgia, while cool tones convey calmness and clarity.  

\item \textbf{Circumplex Model of Affect}~\cite{CMA}.  
Describes emotion along \textbf{Valence} (pleasant–unpleasant) and \textbf{Arousal} (activated–deactivated). Positive, high-arousal states promote preference for vivid, dynamic imagery, whereas calmer moods lead to softer, minimalist visuals.  
\end{itemize}


\noindent\textbf{Personal Context}.\ In addition to psychological attributes, we collect a set of personal context variables, encompassing individual background and daily life factors that may shape aesthetic formation. We collect information on participants’ academic background, which may influence cognitive and aesthetic styles across domains such as Arts, Humanities, STEM, and Business. We also include living styles to capture how lifestyle and identity relate to aesthetic tendencies—for example, artistic individuals often prefer creative or vintage visuals, whereas minimalist users favor clean and balanced compositions. To account for visual exposure and subcultural influence, we record participants’ digital life and interests, including major platforms, content preferences, and hobbies. Finally, clothing and fashion preferences are collected, as dressing style often reflects aesthetic inclinations through preferred colors, textures, and compositions. Together, these personal context variables complement psychological factors to form a holistic profile of user visual preference. Details on the psychological measures, personal context variables, illustrative examples, and the survey design are provided in the Supplementary Material~\cref{sec:profile}$\sim$\ref{sec:stats}.

\subsection{User Profile Collection and Agent Construction}
\label{sec:userdata}

\noindent\textbf{Real-user Profile Collection}.\ We collect demographic and psychological preference data through a 19-item online questionnaire derived from aforementioned profiling schema, including an additional item requesting users to upload their personally preferred images. After quality control and filtering, this stage yields 134 valid responses. The textual responses and uploaded images are then parsed and canonicalized to produce standardized user profiles.

\noindent\textbf{Synthetic-Agent Construction}.\ While the real-user dataset provides high-fidelity preference signals, collecting such data is inherently costly and difficult to scale. Notably, 68 out of 134 users who provided profiling data declined to participate in the subsequent stage requiring final preference selection. Moreover, we observe that the collected user profile distribution is highly skewed, likely due to the targeted outreach of the online survey. We refer readers to the Supplementary Material for additional details. To this end, we introduce a synthetic agent-based construction process that enhances benchmark diversity while enabling scalable training data generation for learning-based methods.

\begin{algorithm}
\caption{Agent Dataset Construction}
\label{alg:agent_construction}
\begin{algorithmic}[1]
\Require Rule set $R = \{(r_i, \text{type}_i)\}_{i=1}^{M}$, where $\text{type}_i \in \{\text{hard}, \text{soft}\}$; thresholds $\tau_{\text{conf}}, \tau_{\text{jac}}$; target size $T$; maximum attempts $A_{\max}$;
\Ensure $D_{\text{agents}}$
\State Initialize $D_{\text{agents}} \leftarrow \emptyset$, $attempts \leftarrow 0$
\While{$|D_{\text{agents}}| < T$ \textbf{and} $\mathit{attempts} < A_{\max}$}
    \State $\mathit{attempts} \leftarrow \mathit{attempts} + 1$; $x \leftarrow \text{Sample}(\text{schema})$; $\mathit{penalty} \leftarrow 0$
    \ForAll{$(r_i, \text{type}_i) \in R$ \textbf{such that} $r_i(x) = \text{False}$}
        \State \textbf{if} $\text{type}_i = \text{soft}$ \textbf{then} $\mathit{penalty} \leftarrow \mathit{penalty} + 1$ \textbf{else continue while}
    \EndFor
    \State $\mathit{sim\_max}(x) \leftarrow \max\limits_{s \in \mathrm{Sig}(D_{\text{agents}})_{\text{last }200}} \frac{|\mathrm{Sig}(x) \cap s|}{|\mathrm{Sig}(x) \cup s|}$
    \If{$(\mathit{penalty} \le \tau_{\text{conf}})$ \textbf{and} $\mathit{sim\_max}(x) \leq \tau_{\text{jac}}$} \State $D_{\text{agents}} \leftarrow D_{\text{agents}} \cup \{x\}$
    \EndIf
\EndWhile
\State Assign IDs, export, and \Return $D_{\text{agents}}$
\end{algorithmic}
\end{algorithm}


To generate scalable, self-consistent synthetic agents, we encode the user profile definitions into a unified schema and generate agent candidates by stochastically sampling from it. As outlined in \cref{alg:agent_construction}, each sampled candidate is validated against a consistency rule set $R$ containing hard and soft rules. Hard rules correspond to logical contradictions that trigger immediate rejection, while soft rules penalize a confidence score. Specifically, the rules check: (A) ratings coherence: penalizing internally inconsistent personal context; (B) value antagonisms: detecting conflicting personal values based on curiosity, rigidity and excitement; (C) cross-domain coherence: penalizing aesthetic or behavioral mismatches across lifestyle and visual preferences; and (D) usage–behavior alignment: penalizing claims of low AI use paired with strong visual consumption patterns.
Finally, candidates are filtered using a Jaccard-based diversity criterion against the signatures of previously accepted agents to ensure population heterogeneity. We refer readers to the Supplementary Material~\cref{sec:agentic}$\sim$\ref{sec:dataexm} for detailed rule implementations, prompt templates, and examples.

\subsection{Benchmark Construction}

\noindent\textbf{Candidate Image Set Generation}.\ Since human implicit visual preferences are intrinsically diverse, highly individualized, and impossible to adequately encapsulate within fixed explicit axes, we design a chain-of-thought prompting strategy that leverages LLMs to reason from user profiles and construct a plausible, multi-dimensional sampling space for each user. Both real and synthetic profiles—spanning disciplinary background, cognitive style, openness, emotional state, and value orientation—are encoded in structured natural-language templates. 
Based on these profiles, our prompt guides the LLM through a three-stage deduction: 
(1) synthesizing demographic and lifestyle attributes to construct a holistic personal narritive; (2) utilizing psychological frameworks (\eg, TIPI, Schwartz Values, EVT, CMA) as associative priors to build a high-dimensional space of implicit cognitive and emotional traits and (3) integrating these naunced signals to generate an unbounded, profile-conditioned sampling space for image generation prompts.
We parse these outputs, remove redundancies, and compile up to 20 enriched, stylistically coherent prompts per user, which are then used to generate a pool of candidate images via a diffusion model like Qwen-Image~\cite{wu2025qwen}.

\noindent\textbf{Preference Image Selection}.\ For each real-user profile, we present the candidate image sets to participants, who select 6–8 images that best match their personal preferences. Participants may also optionally upload additional preferred images. After screening and quality control, this process yields a final set of 76 calibrated user records, each comprising the raw questionnaire responses, standardized profile, and human-selected preferred images. Additional details are provided in the Supplementary Material \cref{sec:survey}. For synthetic agents, the candidate images are automatically ranked and selected based on aesthetic scores.

\subsection{Benchmark Statistics}

For every synthetic agent or real user, we construct standardized evaluation tuples. Given the agent/user's final preference pool, we sample $K \leq 5$ preferred images as the reference set $\{\mathcal{I}_k\}_{k=1}^K$. We then select another image from the same pool to serve as the ground-truth target $\mathcal{Y}_{\mathrm{gt}}$, and employ a Vision-Language Model (VLM) to generate a short caption $\mathcal{X}$ focusing strictly on its primary visual content.
An evaluated model receives $(\mathcal{X},\{\mathcal{I}_k\}_{k=1}^K)$ and must generate an image $\mathcal{Y}$ that aligns with both the prompt and the underlying preferences.

Following our pipeline, the consolidated benchmark comprises $1{,}369$ testcases constructed from $1{,}876$ images collected from $251$ agents/users. The synthetic agent subset contributes $719$ testcases ($1{,}231$ images, $175$ agents), while the real-user subset contributes $650$ testcases ($645$ images, $76$ users).

\section{Experiment}
\label{sec:exp}

\subsection{Problem Setting}

Given a short image generation prompt $\mathcal{X}$ and a set of historical user-preferred images $\{\mathcal{I}_k\}_{k=1}^K$, we seek to generate an output image $\mathcal{Y}$ that is faithful to $\mathcal{X}$ while reflecting the user’s visual preference. Formally, the target conditional distribution can be formulated as
$p_{\theta}\big(\mathcal{Y} \mid \mathcal{X}, \{\mathcal{I}_k\}\big)$

\noindent\textbf{Insights on Problem Formulation}.\ Some existing works approach this problem by designing empirical proxies for the joint conditional distribution. 
In our experiments, we focus on four primary paradigms. First, we consider \emph{test-time tuning} approach like DreamBooth~\cite{ruiz2023dreambooth}, where a user-specific training strategy $f(\theta_u,\{\mathcal{I}_k\})$ is applied so that image sampling proceeds from $p_{f(\theta_u,\{\mathcal{I}_k\})}\!\left(\mathcal{I} \mid \mathcal{X}\right)$. Second, we examine methods that directly generate images using a \emph{joint condition} formulation $p_{\theta}\big(\mathcal{Y} \mid \mathcal{X}, \{\mathcal{I}_k\}\big)$. Third, we evaluate approaches that employ a 
\emph{condition fusion} module to jointly summarize 
$\{\mathcal{I}_k\}$ and $\mathcal{X}$ into a unified conditioning representation 
$q_{\xi}(\mathcal{X}, \{\mathcal{I}_k\})$, and subsequently adopt the generation 
process using $p_{\theta}\!\left(\mathcal{Y} \mid q_{\xi}(\mathcal{X}, 
\{\mathcal{I}_k\})\right)$. Finally, we include \emph{separate condition} approach that generate images based on decoupled pathways $p(\mathcal{Y} \mid \mathcal{X})$ and $p(\mathcal{Y} \mid \{\mathcal{I}_k\})$.


\subsection{Metrics}

To ensure reproducible evaluation of our benchmark, we adopt a hybrid protocol.
Although automated metrics facilitate efficient model development, they often fall short in capturing the subjective nuances of personalized preferences.
Consequently, as a complementary component, we introduce a persona-aware Elo rating system that employs LLMs as proxy judges to better assess the alignment between the generated images, user-specific preferences, and textual prompts.

\noindent\textbf{Automatic Metrics}.\ Following previous work~\cite{xu2025personalized,kim2025draw}, we evaluate instruction fidelity via CLIP text-image similarity (\textbf{CLS-T})~\cite{hessel2021clipscore}, which quantify the semantic alignment between the generated image $\mathcal{Y}$ and the user's short prompt $\mathcal{X}$. To assess how well $\mathcal{Y}$ incorporates visual conditions from the user's reference image set $\{\mathcal{I}_k\}$, we report the average CLIP image-image similarity (\textbf{CLS-R}), DINO similarity (\textbf{DIS-R})~\cite{oquab2023dinov2}, and LPIPS (\textbf{LPIPS-R})~\cite{zhang2018perceptual}. Note that we adopt DINO and LPIPS, as our personalized generation task extends beyond standard style transfer to capture preference-relevant background semantics, structural integrity, and latent aesthetic styles. Higher CLS/DIS and lower LPIPS scores denote superior alignment. For clarity, CLS and DIS are scaled by a factor of 100. We also calculate their grayscale variants in Supplementary Material~\cref{sec:eval_IF}.

\noindent\textbf{Persona-aware Elo Rating}.\ Since real-world user preferences are inherently diverse and non-singleton, relying solely on image-based automated metrics is insufficient. Therefore, we adopt an arena-style~\cite{jiang2024genai} evaluation using \textbf{LLM-as-a-judge} to conduct pairwise comparisons on our Real-User dataset, ensuring reproducible assessment. The judge is provided with the complete \textbf{user profile} and explicitly instructed to act as that specific user. Operating under this assigned persona, the evaluator assesses two generated images and judges which one better aligns with the user's underlying aesthetic preferences.

To ensure fairness and mitigate position bias, we randomly swap the presentation order of the two evaluated images. Furthermore, to avoid self-enhancement bias, we employ multiple independent judge models, including GPT-5~\cite{openai25-gpt}, Gemini 2.5 Pro~\cite{deepmind2025gemin}, and Qwen3-VL~\cite{bai2025qwen3}. Following~\cite{xu2025investigating}, we process each committee member's decision as an independent pairwise evaluation, expressly avoiding the information loss associated with majority voting. To validate the reliability of these persona-aware judging mechanism, we compare a sampled subset of the committee's results against human annotations. This comparison yields an agreement rate of $\sim 91\%$, demonstrating that our auto-judge serves as a highly consistent proxy for reproducible, human-aligned preference evaluation. Based on these fine-grained pairwise judgments, we compute an \textbf{Elo rating} for representative methods to quantify their capacity to capture individual tastes. 

To further verify the reliability of the auto-ELO score, we conduct a user study. 
We select the top four models according to the ELO results in Table~\ref{tab:three_datasets} and sample 100 cases for human evaluation. For each case, human annotators are given the user profile and reference image, and asked to select the best result among the four model outputs. The ranking is broadly aligned with our auto-ELO evaluation, suggesting that the persona-aware auto-judge provides a reasonable proxy for human preference while enabling efficient benchmark-scale evaluation.
We provide implementation details in the Supplementary Material~\cref{sec:arena}.

The automatic metrics and the persona-aware Elo ratings in~\cref{tab:three_datasets} indicate that automatic metrics closely align with their overall Elo rankings, with top-performing models like GPT-5 consistently achieving both the highest Elo ratings and the most best automatic scores. This strong correlation demonstrates that our proposed hybrid evaluation framework provides a highly accurate, reliable, and comprehensive assessment of personalized image generation quality.

\subsection{Analysis of Preference Conditioning Methods}

We denote the baseline Qwen-Image model without any preference conditions as \textbf{no-preference}. For \emph{test-time tuning} approaches, which explicitly adapt the model parameters to a user's specific reference set prior to inference, we evaluate the widely adopted DreamBooth~\cite{ruiz2023dreambooth} framework on Qwen-Image~\cite{wu2025qwen}. For joint conditioning, we evaluate Qwen-Image-Edit~\cite{wu2025qwen} with randomly sampled references from ${\mathcal{I}_k}$. The (1-Ref) variant uses one reference image as guidance, while the (2-Ref) variant uses two references for joint conditioning.

For \emph{condition fusion} methods, we examine different VLMs that enrich the short caption $\mathcal{X}$ based on the reference images $\{\mathcal{I}_k\}$. The expanded description is then fed to the same image generation model Qwen-Image~\cite{wu2025qwen}. Specifically, we consider the following VLMs for condition fusion, including proprietary models such as GPT-5~\cite{openai25-gpt} and Gemini 2.5-Pro~\cite{deepmind2025gemin}, as well as open-source Qwen models~\cite{Qwen2.5-VL} of different scales. For fair comparison, all models adopt the same prompt, which is disclosed in the supplementary material~\cref{sec:train_vlm}. Finally, for \emph{separate condition} modeling, we only evaluate Fabric~\cite{von2024fabric}, as few studies have explored this setting for personalized image generation.

\begin{table}[t]
\footnotesize
\centering
\setlength{\tabcolsep}{2pt}
\caption{Results of representative methods on our PIPBench. Best and second-best scores across conditioning methods are shown in \textbf{bold} and \underline{underlined}, respectively.}
\resizebox{\linewidth}{!}{

\begin{tabular}{lccccccccc}
\toprule
\multirow{2}{*}{Model} & 
\multicolumn{4}{c}{Synthetic Agent} & 
\multicolumn{5}{c}{Real-User} 
\\
\cmidrule(lr){2-5}\cmidrule(lr){6-10}
 & CLS-T $\uparrow$ & LPIPS-R $\downarrow$ & CLS-R $\uparrow$ & DIS-R $\uparrow$ & CLS-T $\uparrow$ & LPIPS-R $\downarrow$ & CLS-R $\uparrow$ & DIS-R$\uparrow$ & Elo$\uparrow$ \\
\midrule
no-preference &  32.461 &  0.7559 & 62.784 & 10.195 & 32.677 &0.7448 & 62.761 & 12.099 & 1427 \\
\hline
\multicolumn{6}{l}{\textit{Test-time Tuning}} \\
DreamBooth & \textbf{31.693} & 0.7374 & 63.444 & 10.705 & \textbf{31.756} & 0.7265 & 63.751 & 12.720 & 1452 \\
\hline
\multicolumn{6}{l}{\textit{Joint Conditioning Models}} \\
Qwen-Image-Edit (1-Ref) &  30.614 & 0.7326 & 64.615 & 15.461 & 31.048 & 0.7209 & 65.802 & 18.459 & 1521\\
Qwen-Image-Edit (2-Ref)&  30.753 & 0.7652 & 62.319 & 12.259 & 30.559 & 0.7511 & 65.023 & 16.821 & 1354\\
\hline
\multicolumn{6}{l}{\textit{VLM Conditioning Fusion}} \\
GPT-5 &  30.724 & \textbf{0.6977} & \textbf{68.119} & \underline{17.085} & 30.550 & \textbf{0.6867} & \textbf{69.574} & \textbf{22.160} & \textbf{1765}\\
Gemini2.5-Pro &  30.621 & \underline{0.7038} & \underline{67.335} & 14.403 & 30.403 & \underline{0.6910} & \underline{69.174} & 20.090 & \underline{1615} \\
QwenVL2.5-7B &  30.496 & 0.7434 & 64.639 & 13.313 & 30.510 & 0.7320 & 65.460 & 17.275 & 1445\\
QwenVL2.5-32B &  30.937 & 0.7376 & 65.252 & 13.589 & 30.915 & 0.7229 & 66.368 & 17.322  & 1477\\
QwenVL2.5-70B &  29.850 & 0.7183 & 65.894 & \textbf{17.336} & 29.797 & 0.7092 & 66.835 & \underline{20.282} & 1531 \\
\hline
\multicolumn{6}{l}{\textit{Separate Conditioning}} \\
Fabric &  \underline{31.131} & 0.7277 & 64.531 & 13.367 & \underline{31.213} & 0.7222 & 64.590 & 15.463 & 1412 \\
\bottomrule
\end{tabular}
}

\label{tab:three_datasets}
\end{table}

\begin{table*}[t]
\footnotesize
\centering
\setlength{\tabcolsep}{2pt}
\caption{Results on our benchmark using different generation backbones. We use GPT-5 as the conditioning fusion method.}
\resizebox{\linewidth}{!}{
\begin{tabular}{lccccccccc}

\toprule
\multirow{2}{*}{Model} & 
\multicolumn{4}{c}{Synthetic Agent} & 
\multicolumn{5}{c}{Real-User} \\
\cmidrule(lr){2-5}\cmidrule(lr){6-10}
 & CLS-T $\uparrow$ & LPIPS-R $\downarrow$ & CLS-R $\uparrow$ & DIS-R $\uparrow$ & CLS-T $\uparrow$ & LPIPS-R $\downarrow$ & CLS-R $\uparrow$ & DIS-R$\uparrow$ & Elo$\uparrow$ \\
\midrule
GPT-5 + Qwen-Image & 30.724 & 0.6977 & 68.119 & 17.085 & 30.550 & 0.6867 & 69.574 & 22.160 & 1603 \\
GPT-5 + Flux        & 30.740 & 0.7189 & 67.520 & 15.880 & 30.287 & 0.7097 & 68.735 & 20.784 & 1503 \\
GPT-5 + SDXL  & 30.510 & 0.7377 & 66.473 & 14.390 & 29.987 & 0.7331 & 67.951 & 18.690 & 1394\\
\bottomrule
\end{tabular}}

\label{tab:diffusion_backbone}
\end{table*}

\begin{figure}[t]
    \centering
    \includegraphics[width=1\linewidth, keepaspectratio]{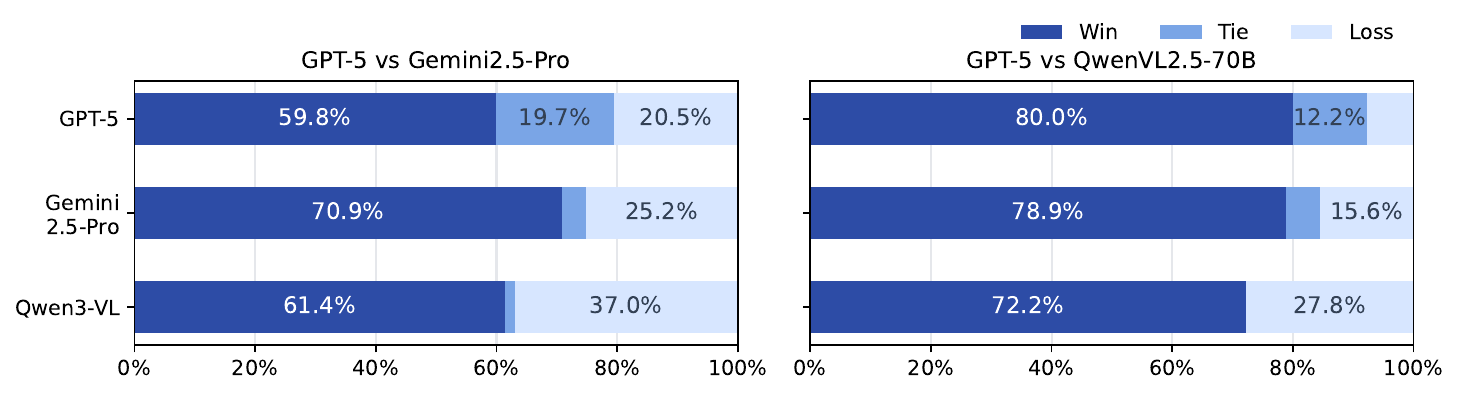}
    \caption{Pairwise preference evaluation results. Win, tie, and loss rates of GPT-5 compared to Gemini 2.5 Pro (left) and QwenVL2.5-70B (right), as evaluated by different VLM judges (Qwen3-VL, Gemini 2.5 Pro, and GPT-5).}
    \label{fig:win_tie_loss}
\end{figure}

\noindent\textbf{The Effect of Preference Conditioning}.\ As shown in~\cref{tab:three_datasets}, most existing methods that incorporate reference images achieve better preference alignment performance compared with the no-preference baseline. Interestingly, results from joint conditional models further show that providing more reference images can actually degrade performance. This is primarily because even state-of-the-art image editing and generation models struggle to jointly interpret multiple reference images, highlighting the need to develop a new class of methods for personalized image generation. Furthermore, it is important to note that high \textbf{CLS-T} scores are not indicative of preference alignment. We therefore use CLIP-based instruction fidelity primarily as a sanity check and a trade-off diagnostic: the ultimate goal is to improve preference alignment while keeping the CLIP drop within a modest range of the no-preference upper bound, avoiding both prompt neglect and uncontrolled over-personalization.

For VLM condition fusion models, proprietary models demonstrate a stronger ability to convert multi-image inputs into coherent textual descriptions that capture users’ visual preferences, thereby yielding superior preference alignment performance. In addition, model scaling shows notable benefits, as evidenced by the results on the Qwen-series VLMs, primarily due to the improved multi-image reasoning capability of larger models.

We note that the separate-condition method, Fabric, is built upon a much weaker image generation backbone. We include this method primarily for benchmarking completeness, and its lower performance should not be interpreted as evidence that separate-condition modeling is ineffective. However, it is important to highlight that separate-condition designs introduce additional computational overhead, making them difficult to adapt to heavy generation backbones.

\noindent\textbf{Comparison Between Synthetic and Real Data}.\ A interesting finding is that aligning preferences for real users is easier than aligning those for synthetic agents. This is because participants in our user study reflect the prevailing preferences commonly seen on the internet, making their visual tastes easier for models to infer. In contrast, the synthetic agents cover a broader and more diverse spectrum of personalities, which makes preference extraction substantially more challenging. These findings highlight the benefits of our data collection strategy: the synthetic and real-user data complement each other under our carefully designed protocol. Moreover, the relative performance gaps among the evaluated methods remain consistent across real and synthetic data, supporting the validity of our synthetic pipeline. Our method thus offers a novel pathway for studying underrepresented social groups. We compare the demographic distribution of collected users and synthetic agents in the Supplementary Material~\cref{sec:stats}.

\noindent\textbf{Analysis of Persona-aware Judges}.\ \cref{fig:win_tie_loss} details the win, tie, and loss distributions produced by each judge during pairwise comparisons of the leading baselines. This analysis yields two key observations. First, self-enhancement bias remains minimal, as judges do not systematically favor models from their own families, benefiting from our carefully designed prompts. Second, we observe distinct evaluation behaviors under persona-conditioned assessment: GPT-5 tends to be more conservative, producing more tie verdicts, whereas Qwen3-VL is more decisive in issuing win–loss judgments. Using three independent judges therefore enhances the robustness of our evaluation. The full arena results are presented in the Supplementary Materials~\cref{sec:arena}.

\noindent\textbf{Results across Different Generation Backbones}.\ To assess how different image generation models affect preference alignment, we evaluate three widely used generation backbones, all paired with the GPT-5–based condition fusion method. The experimental results are presented in ~\cref{tab:diffusion_backbone}. We note that the reference images are generated using Qwen-Image, which gives this model an inherent advantage on automatic metrics.
To mitigate this influence, we concurrently employ the profile-based Elo, providing a fairer comparison of the underlying backbone performances.
As expected, Qwen-Image achieves the best alignment, followed by Flux. Their larger model capacities provide stronger expressiveness for capturing the nuanced textual descriptions of user preferences.
\begin{table*}[t]
\centering
\setlength{\tabcolsep}{2pt}
\footnotesize
\caption{Results on our benchmark using different learnable methods to encode preference image embeddings. For each column, the best score (without no-preference baseline) is \textbf{bold}, and the second-best score is \underline{underlined}.}
\resizebox{\linewidth}{!}{
\begin{tabular}{lccccccccc}
\toprule
\multirow{2}{*}{Model} & 
\multicolumn{4}{c}{Synthetic Agent} & 
\multicolumn{5}{c}{Real-User} \\
\cmidrule(lr){2-5}\cmidrule(lr){6-10}
 & CLS-T $\uparrow$ & LPIPS-R $\downarrow$ & CLS-R $\uparrow$ & DIS-R $\uparrow$ & CLS-T $\uparrow$ & LPIPS-R $\downarrow$ & CLS-R $\uparrow$ & DIS-R$\uparrow$ & Elo$\uparrow$ \\
\midrule
no-preference &  32.461 & 0.7559 & 62.784 & 10.195 & 32.677 & 0.7448 & 62.761 & 12.099  & 1485 \\
QwenVL2.5-7B & 30.496 &  0.7434 & 64.639 & 13.313 & 30.510 & 0.7320 & 65.460 & 17.275 & 1492 \\
QwenVL2.5-32B & 30.937 &  0.7376 & 65.252 & 13.589 & 30.915 & 0.7229 & \underline{66.368} & \underline{17.322} & 1504 \\
\hline
\multicolumn{6}{l}{\textit{VLM Instruction Tuning}} \\
Qwen2.5VL-7B-finetuned &  \underline{31.510} & \textbf{0.7002} & \underline{65.283} & \underline{13.590} & \underline{31.588} & \textbf{0.6901} & 66.350 & 17.013 & \textbf{1597} \\
Qwen2.5VL-32B-finetuned &  31.322 & \underline{0.7081} & 64.834 & \textbf{13.652} & 31.396 & \underline{0.6950} & \textbf{66.814} & \textbf{17.444} & \underline{1570} \\ 
\hline
\multicolumn{6}{l}{\textit{Preference Compression}} \\
Learnable Compressor     &  \textbf{31.716} & 0.7463 & \textbf{65.333} & 12.184 & \textbf{31.994} & 0.7270 & 65.264 & 14.528 & 1534 \\
Frozen Compressor   & 29.409 & 0.7440 & 62.133 & 10.196 & 29.254 & 0.7248 & 63.612 & 11.883 & 1318 \\
\bottomrule
\end{tabular}}
\label{tab:learn}
\end{table*}
\subsection{Optimize Preference Conditioned Generation}

Our agent-based data construction process not only complements benchmark data construction, but also provides an effective mechanism for generating training data used to learn preference embeddings. To recap, each agent is associated with multiple preferred images, and we randomly sample from these images to compose the reference set $\{\mathcal{I}_k\}$ and target image $\mathcal{Y}_{\mathrm{gt}}$. We denote the short caption for testing time input as $\mathcal{X}$. And the expanded caption that originally used to generate $\mathcal{Y}_{\mathrm{gt}}$ is denoted as $\mathcal{E}$. We consider the following learnable approaches designed to optimize the encoding of image preferences:

\noindent\textbf{VLM Instruction Tuning}.\ We fine-tune Qwen2.5-VL-7B and Qwen2.5-VL-32B to generate enriched captions conditioned on both the reference images and the short caption. Concretely, the VLM is optimized to model the joint distribution $p\big(\mathcal{E} \mid \mathcal{X}, \{\mathcal{I}_k\}\big)$. See the Supplementary Materials ~\cref{sec:train_vlm} for details.

\noindent\textbf{Preference Compression}.\ We first employ a VLM-based preference extractor to summarize a user’s reference images into a compact set of preference tokens that encode their visual tastes. A lightweight cross-attention compressor further embed the dense multi-image features into these tokens, which are subsequently injected into the DiT blocks via in-block adapters to guide personalized generation. Since the VLM is already employed for visual reference comparison, we report model performance excluding VLM-based caption enrichment to ensure a fair comparison with respect to computational cost. We refer to the Supplementary Materials~\cref{sec:train_preference} for the exact compressor design and training details.

The experimental results are presented in~\cref{tab:learn}. Visual instruction tuning improves overall performance by a small margin across all three levels of data, even though the training data comes solely from synthetic agents. Particularly, the improvement observed on real-user data further supports the validity of our agent construction pipeline. As for the preference compression methods, we find that freezing the VLM and training only the projection layer is suboptimal for encoding visual  preferences. When we unfreeze the VLM, the model does learn to better attend to the reference images compared to the no-preference. However, this approach remains less effective compared to prompt re-writing method using VLMs. This observation highlights an exciting research direction focused on developing fully end-to-end models for visual preference learning.

\subsection{Analysis on Importance of Profile-inclusive Framework}
\label{subsec:analysis_importance}
To demonstrate the necessity of our proposed profile-conditioned generation paradigm, we compare our synthetic agent pipeline against a \emph{profile-free} baseline. While our framework constructs preference sets by conditioning LLM agents on rich, holistic user profiles, the baseline method generates data based solely on isolated, manually specified aesthetic tags. For a fair comparison, we generate both benchmark and training datasets using this profile-free pipeline at identical scale and comparable complexity to those constructed by our pipeline.

\noindent\textbf{Benchmark Data Quality}.\ By comparing the benchmark datasets generated by the profile-free pipeline and our profile-inclusive pipeline (denoted as Ours), we observe that Ours better reflects the practical complexities of human preferences. Specifically, Ours exhibits lower intra-user similarity (CLIP: \textbf{64.18} vs.\ 65.49; DINO: \textbf{12.35} vs.\ 23.70), indicating that profile-conditioned generation yields more diverse within-user preference sets, rather than aesthetic repetitions. Furthermore, at the inter-user level, Ours demonstrates lower mean Shannon entropy (\textbf{0.5811} vs.\ 0.9919) and higher Silhouette scores for $K\!\in\![2,50]$. These results indicate improved clusterability, suggesting that our method better captures distinct and diverse individual visual preferences as in real-world scenarios. These quantitative findings are corroborated by an user study assessing the diversity, coherence, and consistency of the reference-target pairs on a 1--5 scale, where Ours achieves a superior score of \textbf{3.61} compared to 3.28 for profile-free method. We refer to the Supplementary Material~\cref{sec:details_importance} for additional details.

\begin{table}[t]
\centering
\footnotesize
\caption{Comparison of VLM instruction-tuning performance on real-user data using Qwen2.5-VL-7B trained on data from the profile-free pipeline versus our pipeline.}
\label{tab:training_data_quality}
\small
\begin{tabular}{lccccc}
\toprule
Method & CLS-T$\uparrow$ & LPIPS-R $\downarrow$ & CLS-R $\uparrow$ & DIS-R $\uparrow$ & win rate $\uparrow$\\
\midrule
profile-free & \textbf{32.251} & 0.7192 & 64.596 & 13.574 & 47.6\%\\
\textbf{profile-inclusive (Ours)} & 31.588 & \textbf{0.6901} & \textbf{66.350} & \textbf{17.013} & \textbf{52.4\%}\\
\bottomrule
\end{tabular}
\end{table}

\noindent\textbf{Training Data Quality}.\ We compare VLM instruction-tuning performance on real-user data using Qwen2.5-VL-7B as the base model, trained separately on the datasets generated by the two pipelines. As shown in~\cref{tab:training_data_quality}, the profile-free pipeline lags behind ours by a notable margin across all visual preference alignment metrics (LPIPS-R, CLS-R, DIS-R) and persona-aware judge. This performance drop suggests that synthesizing training data using only explicit aesthetic tags fails to capture the multifaceted complexity and diversity of real users’ visual choices. Note that the explicit model attains a slightly higher text alignment score (CLS-T), likely because it underutilizes visual reference signals and thus behaves closer to a no-preference baseline.


\begin{figure*}[t]
    \centering
    \includegraphics[width=1\linewidth, keepaspectratio]{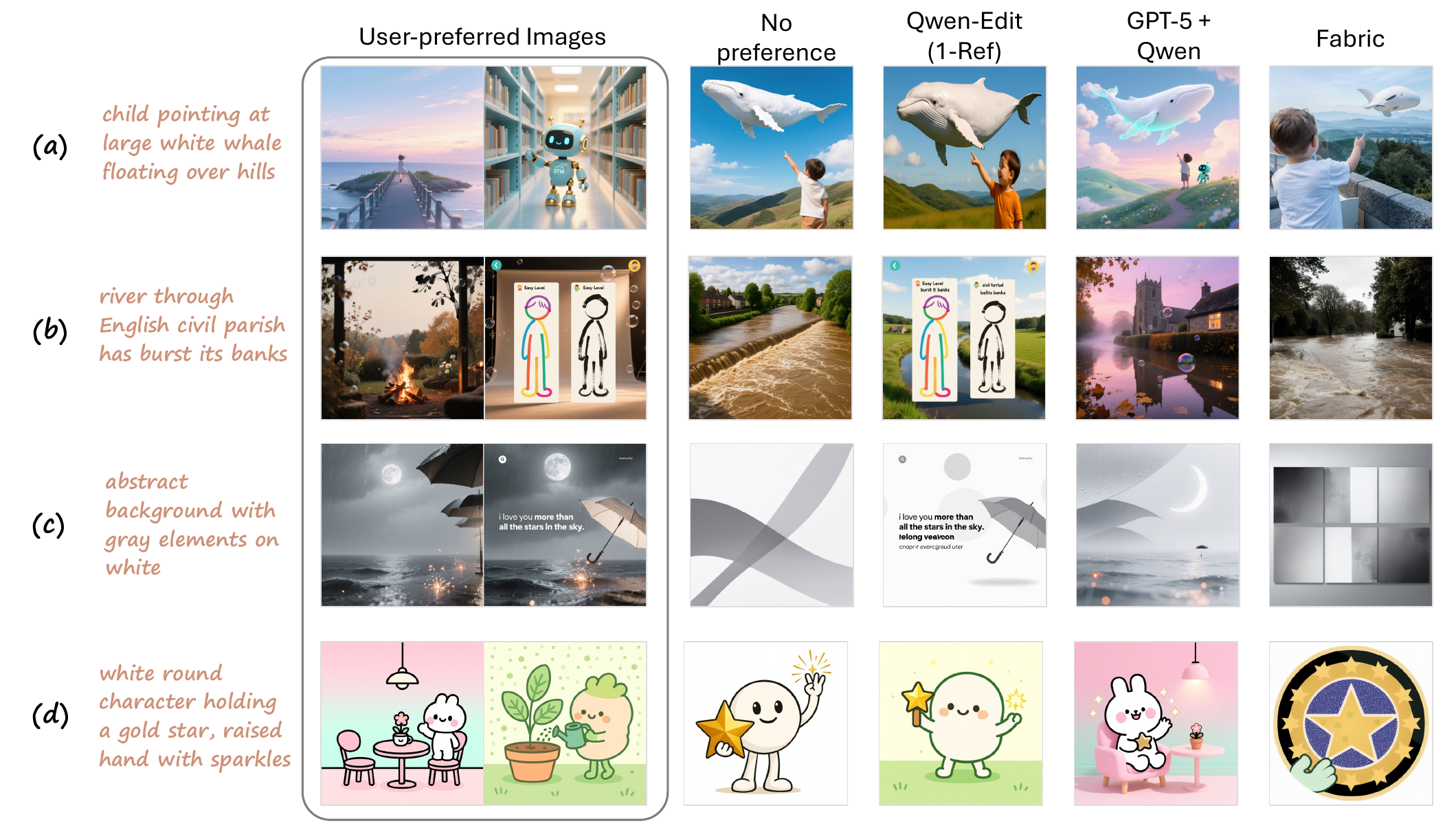}
    \caption{Visualizations of personalized image generation by different methods}
    \label{fig:comparison}
\end{figure*}
\subsection{Visualizations}

We provide visualizations of personalized image generation in~\cref{fig:comparison}, comparing the no-preference baseline, the Qwen-Image-Edit (1-Ref) \emph{joint-condition} method, the GPT-5–based \emph{preference fusion} method, and the Fabric \emph{separate-condition method}.
Panels (a) and (b) are weak cases, while (c) and (d) show stronger alignment with user tastes.
All preference-aware variants add user-style traits beyond the text input; in (c) and (d) these traits more clearly match the preferred images. However, Qwen-Edit (1-Ref), which relies on a single random reference, can be dominated by that image—for example, in (b) it almost copies the second preferred image—indicating that one picture is insufficient to represent user preference. GPT-5–based fusion may also over-emphasize preference: in (a), GPT-5 + Qwen adds a robot seen once in the reference set, contradicting the original prompt and illustrating a conflict between preferences following and instruction fidelity.
\section{Conclusion}
\label{sec:conc}

In this paper, we introduce PIPBench, the first profile-grounded benchmark designed to assess whether existing image generation models can effectively leverage historical user data for preference-aligned image generation. We propose a novel data collection pipeline that enables high-quality user profile collection and the construction of synthetic agents that emulate human personalities in visual preference. We conduct comprehensive experiments to demonstrate the limitations of existing methods in encoding user preferences and to validate the effectiveness of our proposed pipeline. We believe that PIPBench represents an important step toward user-driven image generation models and highlights promising research opportunities in developing end-to-end approaches for visual preference encoding. In addition, our synthetic data engine and user study protocol offer a meaningful attempt at constructing agents that can facilitate a deeper understanding of underrepresented user groups.

\noindent\textbf{Acknowledgments}.\ This work was supported in part by the Zhiyuan Scholar Program of the Beijing Municipal Science and Technology Commission (Z251100008125045), NSFC Grants, and a research grant from the ByteDance Seed Team.

\bibliographystyle{splncs04}
\bibliography{main}

\clearpage
\appendix
\setcounter{page}{1}

\renewcommand\thesection{\Alph {section}}
\renewcommand\thesubsection{\thesection.\arabic{subsection}}

\renewcommand{\theHsection}{appendix.\Alph{section}}
\renewcommand{\theHsubsection}{appendix.\Alph{section}.\arabic{subsection}}

This is the supplementary material for the paper "PIPBench: A Profile-Inclusive Framework for Personalized Image Generation Evaluation". We organize the content as follows:
\\

\noindent\textbf{\hyperref[sec:compare]{\ref{sec:compare}} -- Comparison to Related Benchmark} \\[0.2cm]
\noindent\textbf{\hyperref[sec:arena]{\ref{sec:arena}} -- Details on Persona-aware Elo Rating} \\ [0.2cm]
\noindent\textbf{\hyperref[sec:eval_IF]{\ref{sec:eval_IF}} -- Evaluating Beyond Color-related Preferences} \\[0.2cm]
\noindent\textbf{\hyperref[sec:train_vlm]{\ref{sec:train_vlm}} -- Details on Visual Instruction Tuning} \\ [0.2cm]
\noindent\textbf{\hyperref[sec:train_preference]{\ref{sec:train_preference}} -- Details on Preference Compression} \\ [0.2cm]
\noindent\textbf{\hyperref[sec:details_importance]{\ref{sec:details_importance}} -- Details on Analysis on Importance of Profile-inclusive Framework} \\[0.2cm]
\noindent\textbf{\hyperref[sec:profile]{\ref{sec:profile}} -- Profile Definition} \\[0.2cm]
\noindent\textbf{\hyperref[sec:survey]{\ref{sec:survey}} -- Details on User Survey} \\ [0.2cm]
\noindent\textbf{\hyperref[sec:stats]{\ref{sec:stats}} -- User Survey Statistics} \\ [0.2cm]
\noindent\textbf{\hyperref[sec:agentic]{\ref{sec:agentic}} -- Details on Agentic Data Construction} \\ [0.2cm]
\noindent\textbf{\hyperref[sec:caption]{\ref{sec:caption}} -- Details on Image Caption Generation} \\ [0.2cm]
\noindent\textbf{\hyperref[sec:dataexm]{\ref{sec:dataexm}} -- Data Example} \\ [0.2cm]
\noindent\textbf{\hyperref[sec:vis]{\ref{sec:vis}} -- Additional Visualization} \\[0.2cm]
\noindent\textbf{\hyperref[sec:reproduce]{\ref{sec:reproduce}} -- Reproducibility Statement} \\[0.2cm]
\noindent\textbf{\hyperref[sec:limitations]{\ref{sec:limitations}} -- Limitations and Future Work}

\section{Comparison to Related Benchmark}
\label{sec:compare}

Table~\ref{tab:compare_benchmark} summarizes how PIPBench relates to representative benchmarks for preference-aware or personalized image generation. For works that mainly report method-centric user studies rather than a reusable benchmark (\eg, ViPer, InstaPA), we either approximate the evaluation scale from their user-study setup (ViPer: roughly $2{,}400$ evaluated images and 20 users) or leave the entries as ``--'' when image/user counts are not clearly specified (InstaPA). For DreamBench++, we follow the common practice of reporting the number of subjects/reference images (150), as it personalizes to subjects rather than real users. For TailoredV, PMG, and Pigeon, we report the scale of their released or clearly defined test settings, aggregating over their image-generation scenarios.

Compared to these benchmarks, \textbf{PIPBench} has three key advantages. First, it is the \emph{only} benchmark in Table~\ref{tab:compare_benchmark} that includes explicit real user profiles (demographics, psychological traits, and aesthetic axes), rather than relying solely on historical prompts, interaction logs, or single reference images. Second, it operates in a general-domain text-to-image setting while maintaining realistic scale (1,876 images and 251 users), bridging the gap between small, carefully controlled user studies (\eg, ViPer) and large but domain-specific recommender-style datasets (PMG, Pigeon). Third, PIPBench is explicitly designed around multi-image preference input: each testcase uses $2\!\sim\!5$ preferred images plus a short prompt, aligning closely with real personalized-generation scenarios, whereas most existing benchmarks either use a single reference image or implicit interaction traces without curated preference examples.

\section{Details on Persona-aware Elo Rating}
\label{sec:arena}

\noindent\textbf{Overview}.\
Concretely, we randomly sample roughly one sixth of the Real-User test cases while preserving the distribution of users and prompts. Each selected case provides reference images $\{\mathcal{I}_k\}_{k=1}^K$, a short prompt $\mathcal{X}$, and a ground-truth target image $\mathcal{Y}_{\mathrm{gt}}$.

\noindent\textbf{Persona-aware LLM-as-a-Judge Comparison Details}.\
For every unordered pair of models $(A,B)$ and every test case, both models generate images $\mathcal{Y}_A$ and $\mathcal{Y}_B$ conditioned on the same $(\mathcal{X}, \{\mathcal{I}_k\}_{k=1}^K)$. These two outputs are then compared by a strong LLM-as-a-Judge, which receives the prompt, the complete user profile, and the two candidate generations. The judge first produces aspect-wise assessments, and finally outputs a single verdict indicating which image better matches the user’s preferences or tie, ac onfidence between 0-1, and a detailed reason. We assume that confidence lower than $0.70$ as tie. We randomly flip the order of $(A,B)$ to reduce position bias.

\begin{figure}[t]
    \centering
    \includegraphics[width=1\linewidth, keepaspectratio]{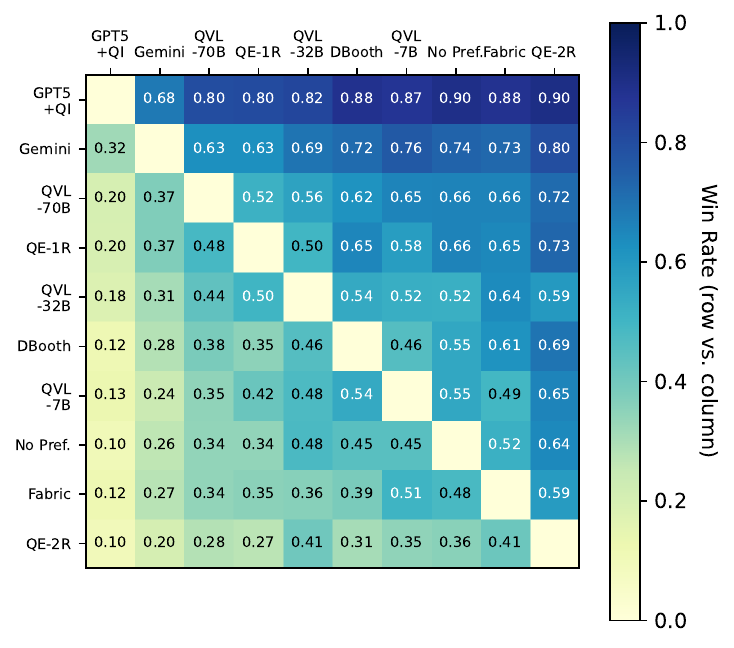}
    \caption{Pairwise win-rate matrix. Each cell shows the probability that the row model is preferred over the column model by the VLM-as-a-Judge, conditioned on the same Real-user profile and prompt; models are ordered by their Elo ratings. 
}
    \label{fig:winrate}
\end{figure}

Aggregating all verdicts yields pairwise win counts between models. While we report the global \emph{Elo ratings} in the main text—providing an overall ranking by fitting a Bradley–Terry style model to these outcomes—we provide a more granular head-to-head analysis in this appendix. Specifically, we report the \emph{pairwise win-loss matrix} in Figure~\ref{fig:winrate}, where the win rate of model $i$ against model $j$ is calculated by counting the clear wins and evenly distributing the ties between both models (i.e., treating each tie as a half-win for both sides). This detailed matrix highlights the specific relative strengths and weaknesses among the evaluated models.

\begin{table*}[t]
\centering
\small
\caption{Comparison of PIPBench with representative benchmarks for preference-aware or personalized image generation. }
\resizebox{\linewidth}{!}{
\begin{tabular}{ccccccc}
\toprule
\textbf{Benchmark} &
\textbf{\# of images} &
\textbf{\# of users} &
\textbf{real user profile} &
\textbf{construction process} &
\textbf{domain} &
\textbf{preference input} \\
\midrule

ViPer~\cite{salehi2024viper} & $\sim$2,400 & 20 & \XSolidBrush & real & general & comment \\
 TailoredV~\cite{chen2024tailored} & 6,230 & 3,135 & \XSolidBrush & synthetic \& real  & general & prompt \\
  InstaPA~\cite{li2025instantpreferencealignmenttexttoimage} & - & - & \XSolidBrush & synthetic \& real & \makecell{culture, art, \\ emotion, movie} & 1 image \\
 DreamBench++~\cite{ICLR2025_71ad539a} & 150 & - & \XSolidBrush & synthetic & general & 1 image\\
 PMG~\cite{xu2025personalized} & 25,100 & 2,600  & \XSolidBrush & synthetic \& real & dialogue & user interaction\\
 Pigeon~\cite{xu2025personalized} & 39,739 & 600 & \XSolidBrush & real & \makecell{sticker, \\movie poster} & images \\
 \textbf{PIPBench (Ours)} & 1,876 & 251 & \Checkmark & synthetic \& real & general & 2 $\sim$ 5 preferred images \\
\bottomrule
\end{tabular}}
\label{tab:compare_benchmark}
\end{table*}


\begin{table}[t]
\centering
\scriptsize
\setlength{\tabcolsep}{4pt}
\renewcommand{\arraystretch}{0.95}
\caption{
Human preference study on 100 sampled cases using the top four models from main paper Table~1.
}
\label{tab:user_study_top4}
\begin{tabular}{lcc}
\toprule
Model & \#Selected & Human Rank \\
\midrule
GPT-5            & \textbf{41} & \textbf{1} \\
Gemini-2.5-Pro   & 22 & 2 \\
Qwen-VL-70B      & 19 & 3 \\
Qwen-Image-Edit  & 18 & 4 \\
\bottomrule
\end{tabular}
\end{table}

\noindent\textbf{Human agreement}.\
To validate the reliability of the VLM-as-a-Judge, we additionally measure its agreement with human annotators on a subset of arena comparisons. We randomly sample $200$ test cases from the results of comparison, balancing over users, prompts, and model pairs. For each sampled case, we present the user’s complete profile, the short prompt $\mathcal{X}$, and two anonymized candidate generations to human raters, with the left--right order of the two candidates randomized. Annotators are asked to decide which image better matches the user's visual preferences.

We then compare the VLM’s verdict against this consensus, reporting the VLM--human agreement rate, i.e., the fraction of comparisons where the VLM and the human majority select the same winner.
The agreement rate between human and VLM is $\sim 91\%$, suggesting that the VLM-as-a-Judge provides a reasonably human-consistent proxy for large-scale preference evaluation.

To further verify the reliability of the auto-ELO score, we conduct a user study. 
We select the top four models according to the ELO results in Table~1 of paper and sample 100 cases for human evaluation. For each case, human annotators are given the user profile and reference image, and asked to select the best result among the four model outputs. We then aggregate the number of selections to obtain a human preference ranking. As shown in Table~\ref{tab:user_study_top4}, GPT-5 receives the most selections, followed by Gemini-2.5-Pro, Qwen-VL-70B, and Qwen-Image-Edit. This ranking is broadly aligned with our auto-ELO evaluation, suggesting that the person-aware auto-judge provides a reasonable proxy for human preference while enabling efficient benchmark-scale evaluation.

\section{Evaluating Beyond Color-related Preferences}
\label{sec:eval_IF}

\begin{table}[t]
\footnotesize
\centering
\setlength{\tabcolsep}{2pt}
\caption{Quantitative comparison using grayscale images to eliminate color-related biases. The metrics (CLS-TG, LPIPS-G, CLS-G, DIS-G) correspond to those in Table~\ref{tab:three_datasets}, computed after applying a grayscale transformation to all images.}
\resizebox{\linewidth}{!}{

\begin{tabular}{lcccccccc}
\toprule
\multirow{2}{*}{Model} & 
\multicolumn{4}{c}{Synthetic Agent} & 
\multicolumn{4}{c}{Real-User} 
\\
\cmidrule(lr){2-5}\cmidrule(lr){6-9}
 & CLS-TG $\uparrow$ & LPIPS-G $\downarrow$ & CLS-G $\uparrow$ & DIS-G $\uparrow$ & CLS-TG $\uparrow$ & LPIPS-G $\downarrow$ & CLS-G $\uparrow$ & DIS-G$\uparrow$ \\
\midrule
no-preference & 31.408 &  0.6498 & 64.813 & 9.697 & 31.502 &0.6441 & 65.908 & 11.749 \\
\hline
\multicolumn{6}{l}{\textit{Test-time Tuning}} \\
DreamBooth & \textbf{30.688} & 0.6477 & 65.634 & 10.220 & \textbf{30.563} & 0.6417 & 67.111 & 12.151  \\
\hline
\multicolumn{6}{l}{\textit{Joint Conditioning Models}} \\
Qwen-Image-Edit (1-Ref) &  29.743 & 0.6380 & 66.805 & 14.960 & \underline{30.000} & 0.6280 & 68.547 & 17.941 \\
Qwen-Image-Edit (2-Ref)&  29.751 & 0.6514 & 65.714 & 11.917 & 29.352 & 0.6432 & 68.308 & 16.485\\
\hline
\multicolumn{6}{l}{\textit{VLM Conditioning Fusion}} \\
GPT-5 &  \underline{30.118} & \textbf{0.6260} & \textbf{69.540} & \underline{16.355} & 29.776 & \underline{0.6150} & \textbf{71.453} & \textbf{21.224}\\
Gemini2.5-Pro &  29.906 & \underline{0.6262} & \underline{69.025} & 13.766 & 29.609 & \textbf{0.6143} & \underline{71.268} & \underline{19.362}  \\
QwenVL2.5-7B &  29.691 & 0.6433 & 66.827 & 12.731 & 29.575 & 0.6334 & 68.471 & 16.476\\
QwenVL2.5-32B & 29.991 & 0.6415 & 67.306 & 12.846 & 29.808 & 0.6309 & 69.054 & 16.548 \\
QwenVL2.5-70B &  29.170 & 0.6361 & 67.697 & \textbf{16.460} & 28.950 & 0.6292 & 69.574 & 19.313  \\
\hline
\multicolumn{6}{l}{\textit{Separate Conditioning}} \\
Fabric &  29.965 & 0.6479 & 66.974 & 12.921 & 29.836 & 0.6437 & 67.758 & 14.786  \\
\bottomrule
\end{tabular}
}

\label{tab:gray_variants}
\end{table}

A potential concern when evaluating personalized image generation is that a benchmark might disproportionately reward models that merely capture superficial, easily learnable color-related styles, such as global tone, lighting, or hue. To demonstrate that our benchmark fundamentally evaluates a much more comprehensive spectrum of user preferences including structural, compositional, and semantic elements, we conducted an additional analysis by deliberately removing all color information.

Building upon the primary evaluation presented in Table~\ref{tab:three_datasets} of the main paper, we converted all reference sets and generated target images to grayscale prior to computing the evaluation metrics. We denote these grayscale-computed metrics as CLS-TG, LPIPS-G, CLS-G, and DIS-G. 

As reported in Table~\ref{tab:gray_variants}, while the absolute numerical values naturally shift due to the absence of color channels, the relative rankings among the evaluated methods and datasets remain entirely consistent with the original RGB-based results. This stable ranking strongly indicates that the stylistic diversity captured by our benchmark is robust and deeply rooted in complex, implicit visual signatures. It confirms that our evaluation protocol effectively measures multi-dimensional personal tastes rather than being dominated by easily observable, explicit color biases.

\section{Details on Visual Instruction Tuning}
\label{sec:train_vlm}

\noindent\textbf{Goal and notation}.\ In addition to image-based preference compression, we train a vision--language model (VLM) to produce \emph{preference-aware enriched captions} conditioned on both the reference images and a short base prompt.
Given a short user prompt $\mathcal{X}$ and a set of reference images $\{\mathcal{I}_k\}_{k=1}^K$, the VLM is optimized to model the conditional distribution
\[
p_{\phi}\big(\mathcal{E} \mid \mathcal{X}, \{\mathcal{I}_k\}\big),
\]
where $\mathcal{E}$ denotes the enriched prompt that adds user-specific aesthetic and semantic cues inferred from $\{\mathcal{I}_k\}$.

\noindent\textbf{Data construction}.\ We construct a dedicated instruction-tuning corpus by reusing the same generation pipelines as for the L1 and synthetic datasets, but sampling from disjoint image pools and user profiles so that the training data remains orthogonal to PIPBench.
Each training example corresponds to a triplet
\[
\big(\mathcal{X}, \{\mathcal{I}_k\}_{k=1}^K, \mathcal{E}_{\mathrm{gt}}\big),
\]
where $\mathcal{X}$ is the short caption (either a filtered LAION-AES caption or a VLM-generated content-only summary), $\{\mathcal{I}_k\}$ are the reference images that characterize a particular user's visual taste, and $\mathcal{E}_{\mathrm{gt}}$ is the enriched prompt that was actually used to generate the ground-truth image in our data pipeline.
We collect approximately $23\text{k}$ such triplets.
To avoid leakage, we ensure that (i) the images involved in this corpus do not appear in PIPBench, (ii) the preference profiles used for constructing $\{\mathcal{I}_k\}$ are disjoint from the benchmark agents/participants, and (iii) the short prompts $\mathcal{X}$ are sampled from non-overlapping caption pools.

\noindent\textbf{Prompt template}.\ For each training example, we wrap the inputs into a fixed instruction template.
The VLM receives the reference images $\{\mathcal{I}_k\}$ as visual inputs, together with a system message and a user message of the following form (line breaks added for clarity):

{\small\ttfamily\raggedright\sloppy
\noindent\textbf{System:} You are a preference inference model. Given several images representing a user's visual tastes, infer the user's aesthetic and semantic preferences (\eg, colors, composition, subjects, styles). When a new text prompt is provided, enrich it with additional details that align with these inferred preferences to make the output more personally fitting. Only output the enhanced prompt.\par

\medskip
\noindent\textbf{User:} [images]\par
Please enhance the following prompt in English:\par
[short prompt].
}

Here, \texttt{[images]} denotes the multi-image input slot populated by $\{\mathcal{I}_k\}$ and \texttt{[short prompt]} is replaced by $\mathcal{X}$.
The target output of the model is the corresponding enriched prompt $\mathcal{E}_{\mathrm{gt}}$.
We use the same template for both Qwen2.5-VL-7B and Qwen2.5-VL-32B, and train separate variants for each model size.


\noindent\textbf{Implementation details}.\ We initialize from the publicly available Qwen2.5-VL-7B and Qwen2.5-VL-32B checkpoints and fine-tune them with the above instruction-tuning objective on 8 x A100 GPUs in parallel.
We adopt a conventional setup with mixed-precision training, gradient clipping, and an AdamW optimizer with a warmup phase followed by a decaying learning-rate schedule.
The core architecture of Qwen2.5-VL is kept unchanged; we only adapt the parameters involved in text generation and vision--language fusion.

\section{Details on Preference Compression}
\label{sec:train_preference}

\noindent\textbf{Overview}.\ As described in the main paper, we summarize a user’s reference images $\{\mathcal{I}_k\}_{k=1}^K$ into a compact set of preference tokens $\mathcal{P} \in \mathbb{R}^{M \times d_{\text{pref}}}$, which are then injected into the DiT backbone as an additional conditioning signal. This section provides architectural and training details of the VLM-based extractor and the cross-attention compressor used to obtain $\mathcal{P}$.

\noindent\textbf{Multi-image encoding with a VLM}.\ We use a pretrained vision--language model (VLM) as a multi-image encoder that contextualizes all reference images jointly.
Given a short system prompt $\mathcal{T}_{\text{sys}}$ (\eg, a fixed instruction for reasoning about user visual preferences) and the image set $\{\mathcal{I}_k\}_{k=1}^K$, we form a multi-modal sequence
\[
\mathcal{S} = [\mathcal{T}_{\text{sys}};\,\mathcal{I}_1;\ldots;\mathcal{I}_K].
\]
The VLM processes $\mathcal{S}$ and produces last-layer hidden states for all visual tokens,
\[
\mathcal{H} = \Phi_{\text{VLM}}(\mathcal{S}) \in \mathbb{R}^{V \times d_{\text{vl}}},
\]
where $V$ is the total number of visual tokens across all reference images and $d_{\text{vl}}$ is the VLM hidden dimensionality.
Because all images are encoded in a single pass with shared attention, $\mathcal{H}$ already captures cross-image correlations such as recurring colors, compositions, and subjects.

In our implementation, $\Phi_{\text{VLM}}$ is instantiated by Qwen2.5-VL-7B.
We fine-tune this model using LoRA adapters (rank $r{=}16$, $\alpha{=}16$, dropout $0.05$) on selected attention and MLP layers, leaving the original pretrained weights frozen to preserve the generic visual and language understanding capabilities.

\noindent\textbf{Cross-attention compressor}.\ To obtain a fixed-size representation independent of the number of reference images and tokens, we map $\mathcal{H}$ to $M$ \emph{preference tokens} using a lightweight cross-attention compressor.
The compressor maintains $M$ learnable query vectors $Q \in \mathbb{R}^{M \times d_q}$ and applies a single cross-attention layer with $\mathcal{H}$ as keys and values:
\[
\mathcal{P} = f_{\text{comp}}(\mathcal{H}) \in \mathbb{R}^{M \times d_{\text{pref}}}.
\]
Internally, $f_{\text{comp}}$ is implemented as standard multi-head attention followed by a linear projection and normalization layer.
Intuitively, each query in $Q$ learns to focus on a different aspect of the user’s visual taste (\eg, color palette, level of abstraction, typical backgrounds), and the resulting tokens in $\mathcal{P}$ serve as a compact, differentiable summary of the reference set.
We use $M{=}32$ preference tokens, set $d_q = d_{\text{vl}}$, and project them to $d_{\text{pref}}$, which matches the hidden size of the DiT image stream.

\noindent\textbf{Integration with the DiT backbone}.\ The preference tokens $\mathcal{P}$ are injected into each DiT block via an in-block adapter that operates on the image token sequence.
Concretely, before the standard text cross-attention, we apply one cross-attention layer where image tokens query $\mathcal{P}$, followed by a gated residual connection back to the image stream.
A timestep-dependent MLP predicts scale/shift/gate parameters that modulate this preference context, so that the influence of $\mathcal{P}$ can vary smoothly across the diffusion trajectory.
The text encoder, VAE, and DiT backbone (including its original image--text attention) remain frozen; only the preference adapter and the compressor are trained, together with LoRA parameters in the VLM.

\noindent\textbf{Training objective and optimization}.\ Training uses the same flow-matching objective as the main model.
For each triplet $(\mathcal{X}, \{\mathcal{I}_k\}, \mathcal{Y}_{\mathrm{gt}})$, we encode $\mathcal{Y}_{\mathrm{gt}}$ into VAE latents $x_0$ and sample Gaussian noise $x_1 \sim \mathcal{N}(0, \mathbf{I})$.
Following Rectified Flow, we sample a timestep $t \in [0,1]$ and construct an interpolated state and velocity:
\[
x_t = (1-t) x_0 + t x_1, \qquad v_t = x_1 - x_0.
\]
The DiT backbone, augmented with the preference adapter, predicts $v_{\theta}(x_t, \mathcal{X}, \mathcal{P}, t)$, and we minimize a time-weighted mean-squared error
\[
\mathcal{L} = \mathbb{E}_{(\mathcal{X},\{\mathcal{I}_k\},\mathcal{Y}_{\mathrm{gt}}), t}\big[\, w(t)\,\| v_{\theta}(x_t, \mathcal{X}, \mathcal{P}, t) - v_t \|_2^2 \big],
\]
where $w(t)$ is a mid-peaked weighting function that emphasizes informative intermediate timesteps.
Preference tokens $\mathcal{P}$ are extracted once per reference set and reused across all timesteps for that sample, amortizing the VLM and compressor cost over the diffusion steps.

\noindent\textbf{Excluding caption enrichment}.\ Although the same VLM could, in principle, be used to enrich text prompts with preference-aware descriptions, we disable such caption enrichment in all reported quantitative experiments.
This choice ensures that comparisons focus on the effect of image-based preference compression and that additional VLM usage does not introduce unfair computational advantages for our method.
Qualitative examples with optional caption enrichment are provided only for visualization and are clearly separated from the main evaluation.

\section{Details on Analysis on Importance of Profile-inclusive Framework}
\label{sec:details_importance}
\noindent\textbf{Details on user study}.\ To rigorously assess the quality and validity of our benchmark dataset, we conducted a score-based user study. Each test case was independently reviewed by two human annotators, who were explicitly instructed to adopt a strict, conservative evaluation standard to ensure high precision and a low tolerance for degenerate preference sets. For every testcase, the annotators were presented with reference image set, the candidate target image, and a brief task prompt. Their objective was to assign a holistic High Set Quality (HSQ) score on a 5-point scale.

The human annotators were asked to evaluate the sets based on three primary dimensions jointly:
\begin{itemize}
    \item \textbf{Reference diversity-with-coherence:} Annotators visually assessed whether the reference set captured meaningful diversity (\eg, varying scenes, contexts, or subjects) while preserving a distinctly stable and recognizable personal taste signature (i.e., ``diverse but still clearly one person'').
    \item \textbf{Reference--target consistency:} Annotators judged whether the target image plausibly belonged to the preference distribution established by the references. This required identifying strong alignment in both semantic content and underlying visual/aesthetic cues.
    \item \textbf{Set realism:} Annotators evaluated the naturalness of the reference collection, ensuring it resembled a genuine, human-curated taste profile rather than an artificial, templated, or synthetically clustered dataset.
\end{itemize}

To prevent score inflation and maintain stringent quality control, the annotators applied a strict penalization framework. They were instructed to aggressively penalize or cap scores upon observing specific failure modes: \textbf{Style/Tag Collapse} (treating sets with over-concentrated aesthetic modes as low-diversity despite differing subjects), \textbf{Template Redundancy} (capping scores at 3 for excessive repetition in framing, background, or mood), \textbf{Mixed-Person Sets} (capping scores at 2 for disjointed sets lacking a coherent signature), and \textbf{Target Mismatch} (capping scores at 3 for targets that only shared superficial traits, such as color tone, without holistic alignment).

The 5-point rubric was deliberately calibrated to be demanding. A perfect score of 5 was reserved exclusively for unambiguous cases demonstrating distinct multi-context diversity, strong target alignment, and natural realism. Lower scores (1 or 2) were assigned to noisy, disjointed collections or cases with severe reference--target discrepancies. The independent scores and rationales from the two human annotators were then collected to determine the final quality assessment for each set.

\begin{table}[t]
\centering
\scriptsize
\setlength{\tabcolsep}{3.5pt}
\renewcommand{\arraystretch}{0.95}
\caption{
Benchmark data quality for different profile settings.
Silhouette scores are reported under different numbers of clusters $K$.
}
\label{tab:preference_diversity}
\begin{tabular}{lcccccc}
\toprule
\multirow{2}{*}{Dataset} 
& \multirow{2}{*}{Shannon Entropy} 
& \multicolumn{4}{c}{Silhouette Score $\uparrow$} & \multirow{2}{*}{Human} \\
\cmidrule(lr){3-6}
& & $K{=}2$ & $K{=}3$ & $K{=}4$ & Avg. \\
\midrule
No-profile    & 0.992 & 0.018 & 0.016 & 0.017 & 0.017& 3.28 \\
Openness-ablated & 0.712 & 0.081 & 0.081 & 0.066 & 0.076 & 3.50\\
\textbf{Full-Profile(Ours)}             & 0.581 & \textbf{0.127} & \textbf{0.109} & \textbf{0.103} & \textbf{0.113} & \textbf{3.61}\\
\bottomrule
\end{tabular}
\end{table}

\noindent\textbf{Ablation on Psychological Profiling}.\ To validate the effecitiveness of the psychological profiling variables, we do a ablation on Openness variable.
Here, we remove all Openness-related traits while freezing other profiling variables and the generation protocol, and regenerate the synthetic data. Following Sec.~\ref{subsec:analysis_importance}, we compute Shannon entropy, Silhouette scores, and conduct the same user-study protocol.
Table~\ref{tab:preference_diversity} shows that Full-Profile achieves the best overall results, with lower entropy, higher Silhouette scores, and better user-study scores. These results indicate that Full-Profile induces more coherent and better-separated preference modes. Removing Openness weakens the cluster structure and user-study performance, though it still outperforms the no-profile variant, supporting the non-trivial role of Openness in structured visual preference modeling.


\section{Profile Definition}
\label{sec:profile}
To evaluate personalized image generation, we require a structured way to represent users’ visual tastes beyond surface-level prompts or tags. Psychological traits, emotional states, color associations, and personal context all provide complementary signals that shape how individuals perceive and prefer visual content. Together, these dimensions form a compact yet expressive profile representation. We now introduce the specific components used in our framework.
\subsection{Big Five}

\noindent\textbf{Definition}.\ The Big Five describes personality along five continuous traits: Openness, Conscientiousness, Extraversion, Agreeableness, and Neuroticism.~\cite{Big5} In this framework, \textbf{Openness} is the most relevant to visual preference.~\cite{Openness}

\noindent\textbf{Relation to aesthetic preference}.\ Openness reflects imagination, aesthetic sensitivity, and receptivity to novelty and variety. Individuals high in Openness tend to enjoy visual art more and engage more with galleries, tolerate ambiguity and multivalence, and welcome unconventional ideas and styles such as modernism or Cubism. They often prefer non-literal or abstract compositions. In contrast, those with low Openness lean toward representational and traditional artworks, with clear subjects and familiar, balanced compositions.~\cite{Openness} 
For example, a high-Openness viewer may favor an asymmetric surreal collage with suggestive symbolism, while a low-Openness viewer may prefer a realistic landscape that conveys order and clarity.

\subsection{Ten-Item Personality Inventory}
\noindent\textbf{Definition}.\ 
The \textbf{Ten-Item Personality Inventory (TIPI)} is an ultra-short personality measure designed for situations where time or survey space is limited. It assesses the Big Five traits using only 10 short statements~\cite{TIPI}, each rated on a seven-point Likert scale (1 = Strongly disagree, 7 = Strongly agree). Each trait is measured by two items -- one positively keyed and one reverse-keyed -- to provide a rapid approximation of personality structure.~\cite{TIPI} \\

\noindent
\textbf{Items used in this framework.} Only \textbf{Items 5 and 10} of the TIPI are utilized, as they specifically capture the Openness dimension:
\begin{quote}
\textit{Item 5: ``I see myself as open to new experiences, complex.''} \\
\textit{Item 10: ``I see myself as conventional, uncreative.'' (reverse-keyed)}
\end{quote}
Higher combined scores on these two items indicate greater Openness to experience.

\subsection{Schwartz’s Basic Human Values}

\noindent\textbf{Definition}.\ 
Schwartz’s model identifies ten universal values (Self-Direction, Stimulation, Hedonism, Achievement, Power, Security, Conformity, Tradition, Benevolence, and Universalism) that guide human motivation and behavior across cultures.~\cite{Schwartz}

\noindent\textbf{Relation to aesthetic preference}.\ 
These values serve as stable thematic anchors influencing artistic taste. High \textit{Self-Direction} or \textit{Stimulation} aligns with innovative and exploratory aesthetics, such as unconventional framing or dynamic compositions. In contrast, high \textit{Tradition} or \textit{Security} values correspond to conservative, classical, and familiar aesthetics characterized by symmetry and predictability. 
For instance, someone valuing \textit{Benevolence} and \textit{Universalism} might prefer tranquil nature scenes that convey harmony between people and the environment, whereas a person high in \textit{Achievement} or \textit{Power} may gravitate toward editorial imagery that emphasizes refinement, status, and control.

\subsection{Ecological Valence Theory}

\noindent\textbf{Definition}.\ 
Ecological Valence Theory (EVT) posits that color preferences arise from the aggregated affective associations people have with objects or events typically linked to each color. In other words, people favor colors that are associated with positive experiences or valued objects.~\cite{EVT}

\noindent\textbf{Relation to aesthetic preference}.\ 
EVT explains how emotional associations with color influence visual taste and stylistic preference.~\cite{EVT}
Warm palettes are often perceived as cozy, nostalgic, or uplifting, while cool palettes evoke freshness, calmness, and trust. As these associations are culturally and experientially grounded, color preference becomes a powerful predictor of visual mood. For example, a person choosing ocean-themed palettes may respond positively to images with soft highlights, open compositions, and gentle contrast—without requiring the image to be predominantly blue.

\subsection{Circumplex Model of Affect}
\noindent\textbf{Definition}.\ 
The Circumplex Model of Affect organizes emotional experience along two orthogonal dimensions: Valence (pleasant–unpleasant) and Arousal (activated–deactivated).~\cite{CMA}

\noindent\textbf{Relation to Aesthetic Preference}.\ 
Short-term emotional states can significantly influence momentary aesthetic choices. A high-valence, high-arousal state tends to make users prefer brighter, more saturated, dynamic, and high-contrast images (\eg, vivid color grading, strong lighting contrast, a sense of motion, or dramatic composition). In contrast, low-valence or low-arousal states are often associated with a preference for softer, less saturated, low-contrast, and calming visual styles (\eg, diffused lighting, foggy atmosphere, and minimalist compositions).

\subsection{Personal Context}
Additionally, we collect a number of background variables that are closely related to aesthetic formation, measured via single-choice or multiple-choice items: 
\begin{itemize}
    \item  Academic background: the user’s field of academic training may influence their cognitive and aesthetic style; we broadly categorize this as Arts and Design, Humanities and Social Sciences, STEM/Medicine, Business/Economics, etc., to analyze possible differences in visual decoding across disciplines.
    \item  Campus role and living style: by asking about the user’s primary roles in campus or community (\eg, academic-oriented, student organization leader, artistic soul, athletic, easygoing) and their living-space style (\eg, cozy and healing, minimalist and functional, e-sports/tech-savvy, vintage/artsy), we capture how lifestyle and identity relate to aesthetic tendencies — for example, “artistic soul” users may prefer vintage or creative visuals, whereas users with a “minimalist” living style tend to favor negative space and simple compositions.
    \item  Digital life and interests: we record frequently used platforms (Instagram, Xiaohongshu, Bilibili, Zhihu, etc.), primary content consumption (life vlogs, film/anime reviews, game streaming, fashion and beauty, popular science, etc.), and core hobbies (film, anime, gaming, photography, reading, visual arts, etc.), since long-term visual exposure and subcultural affiliation can shape visual literacy and preferences.
    \item  Clothing and fashion preferences: we collect preferred dressing styles (\eg, athleisure, minimalist/smart casual, streetwear, darkwear/techwear, sweet/Kawaii), as these preferences often manifest in the depiction of characters’ clothing, color schemes, and props.
\end{itemize}


\section{Details on User Survey}
\label{sec:survey}

\subsection{User Study on Real User Data Collection}

\begin{figure}
    \centering
    \includegraphics[width=1\linewidth, keepaspectratio]{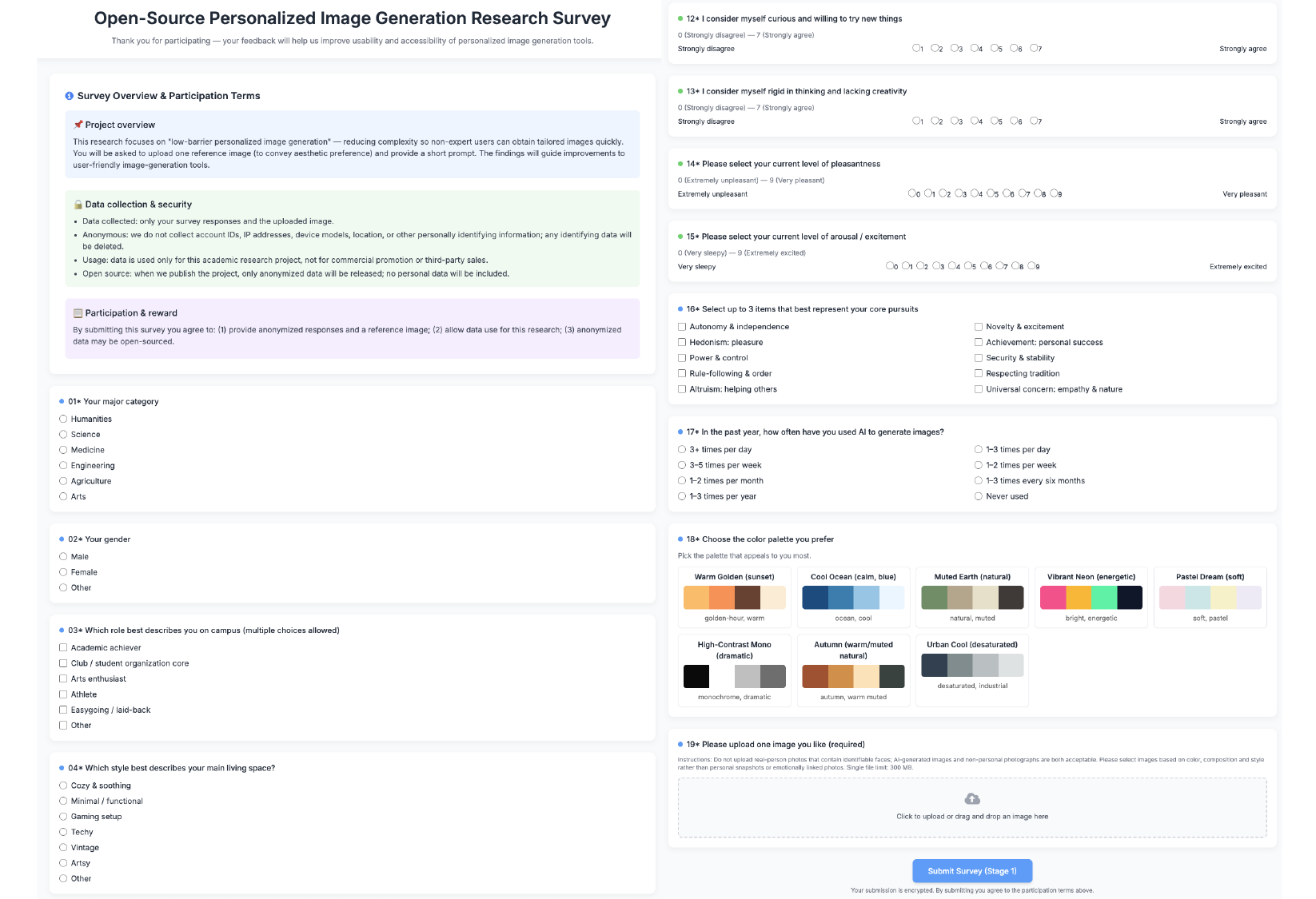}
    \caption{Screenshot of the Stage-1 questionnaire interface, where participants provide demographic and campus background information, report key psychological traits and AI-usage habits, choose representative color palettes, and upload one preferred reference image.}
    \label{fig:website1}
\end{figure}

To construct the Real-user dataset, we designed and carried out a two-stage survey. In the first stage, participant information was collected uniformly via an online questionnaire platform; in the second stage, participants from the first stage were contacted individually to obtain additional data on their visual preferences.

\noindent\textbf{The first stage}.\
In the first stage, based on preliminary research that identified explicit labels and personality traits, we designed an 18-question questionnaire that covers both explicit visual preferences and personality attributes so as to analyze their relationships with visual preference. Specifically, We have designed 10 questions to investigate users' explicit personal characteristics, 8 questions to explore their personality traits, and in the last question, we ask users to upload one or more pictures they like to serve as the basis for subsequent visual analysis. 

For the psychological indicators covered in Section \ref{sec:profile}, given that standard psychological assessment tools contain a large number of items and are relatively time-consuming to administer, this study adopted streamlined psychological measurement methods with acceptable reliability in the questionnaire design. This approach aims to ensure data quality while improving research efficiency. Specifically:
\begin{itemize}
    \item MBTI Personality: A direct self-report format was adopted, inviting users to fill in their known MBTI personality type.
    \item Big Five Personality - Openness Dimension: Questions 5 ("I consider myself open to new experiences and complex in thinking") and 10 ("I consider myself traditional and lacking in creativity") from the TIPI (Ten-Item Personality Inventory) were selected to form a concise assessment module, with a 7-point Likert scale used for the assessment process.
    \item Emotional State: Based on the Circumplex Model of Affect, real-time emotional data collection was conducted through two dimensions: "your current level of emotional pleasure" and "level of emotional arousal".
    \item Schwartz's Theory of Basic Human Values,: In accordance with Schwartz's Theory of Basic Human Values, participants were invited to select the option that best represents their core pursuit from ten categories of universal values, so as to achieve rapid classification.
\end{itemize}

Through the above design, we effectively obtained key psychological characteristic variables while controlling the length of the questionnaire. Table \ref{tab:survey} shows two questionnaire examples, except for Question 19. Finally, a total of 134 complete questionnaires were collected through online recruitment channels. Through online recruitment, 134 participants completed the questionnaire. We present our questionnaires interface in Fig.~\ref{fig:website1}.

\begin{figure}
    \centering
    \includegraphics[width=1\linewidth, keepaspectratio]{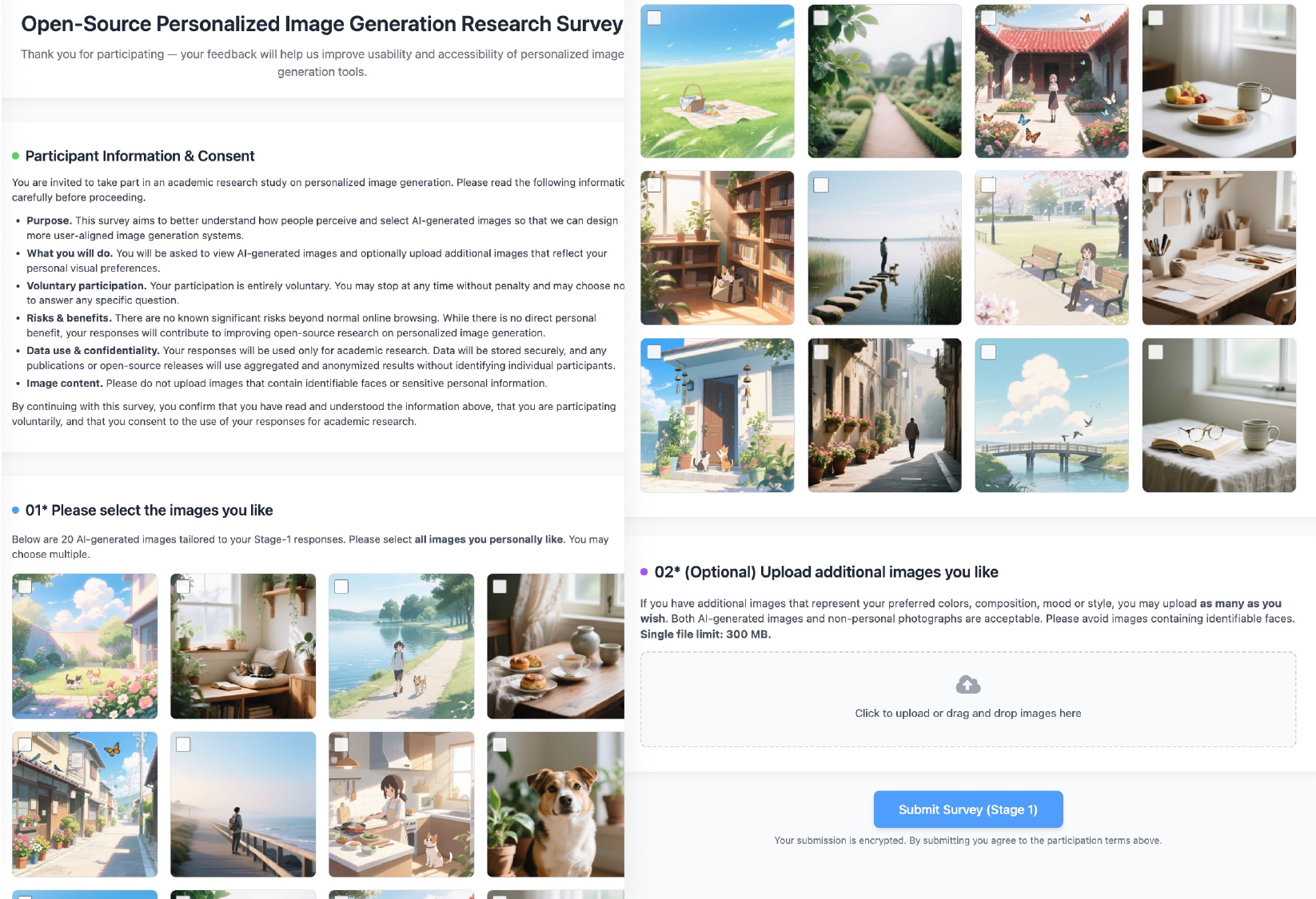}
    \caption{Screenshot of the Stage-2 annotation interface, where participants confirm consent, select their favorite images from their personalized gallery of generated candidates, and optionally upload additional images that reflect their visual preferences.}
    \label{fig:website2}
\end{figure}

To ensure data quality and ethical compliance, we applied multiple quality-control and privacy-protection measures: participants could only proceed to the questionnaire after reading and consenting to an informed-consent form that specified the data usage and authorization; we excluded abnormal responses with completion times significantly below a threshold; we performed manual review and filtering of generated content that involved sensitive or disallowed material; and the released dataset is published only under research IDs. After the above screening, 124 valid questionnaires were retained.

For each valid questionnaire, we invoked the GPT-5 API (hereafter GPT-API) to perform structured analysis of the text responses and the uploaded images. The GPT-API outputs consist of (1) a standardized user profile and (2) 20 diverse image-generation prompts automatically produced from the profile and the uploaded exemplar image. When producing prompts, we deliberately fused questionnaire responses with image preferences — for example, generating brighter, more dynamic descriptions for extraverted users and more tranquil, low-saturation visual descriptions for introverted or contemplative users. We then used an open-source diffusion model to generate, in batch, 20 candidate images per user based on the generated prompts. All generated images underwent manual review to ensure content compliance.

\begin{table*}[t]
  \centering
  \footnotesize  
  \setlength{\tabcolsep}{6pt}    
  \renewcommand{\arraystretch}{1.08}
  \caption{Summary of the questionnaire responses (example)}
  \label{tab:survey}
  \resizebox{\linewidth}{!}{
  \begin{tabular}{@{} c c c @{}}
    \toprule
    Question & User 1 & User 2 \\
    \midrule
    1. Major field of study & Medicine & Science \\
    2. Gender & Female & Male \\
    3. Main role on campus
      & \makecell[c]{“Laid-back” player}
      & \makecell[c]{Core member of a student organization \\ and a “laid-back” player} \\
    4. Which style best describes your living space?
      & Minimalist / functional style & Cozy / healing style \\
    5. Preferred travel type
      & \makecell[c]{Urban exploration, exhibitions}
      & \makecell[c]{Natural scenery, historical sites, \\ cultural tours} \\
    6. Preferred clothing style
      & Simple style & Sporty casual, simple style \\
    7. Hobbies and interests
      & \makecell[c]{Anime, comics, gaming, music, reading, \\ interacting with pets}
      & \makecell[c]{Gaming, handicrafts, exploring nature, \\ interacting with pets} \\
    8. Main content you consume or follow
      & \makecell[c]{Fashion \& beauty, pets, food}
      & \makecell[c]{Anime commentary, game streaming, \\ pets, food} \\
    9. Apps or platforms frequently used
      & \makecell[c]{bilibili, Xiaohongshu, WeChat, Douyin}
      & \makecell[c]{bilibili, Xiaohongshu, Zhihu, \\ WeChat, QQ, Douyin} \\
    10. MBTI type & INTP & ISFJ \\
    11. Preferred thinking style & Rating: 4 & Rating: 4 \\
    12. "I see myself as curious and open to new experiences." & Rating: 7 & Rating: 3 \\
    13. "I see myself as rigid in thinking and lacking creativity." & Rating: 2 & Rating: 4 \\
    14. Current level of pleasant mood & Rating: 5 & Rating: 4 \\
    15. Current level of emotional excitement & Rating: 3 & Rating: 0 \\
    16. Core personal pursuits (choose 3)
      & \makecell[c]{Independence and creativity \\ Novelty and excitement \\ Hedonism: enjoyment}
      & \makecell[c]{Hedonism: enjoyment \\ Security and stability} \\
    17. Frequency of using AI image generation in the past year
      & Never used & 1--3 times a year \\
    18. Preferred color palette & Urban minimalist tones & Cool ocean tones \\
    \bottomrule
  \end{tabular}}
\end{table*}

\begin{figure}
    \centering
    \includegraphics[width=\linewidth, keepaspectratio]{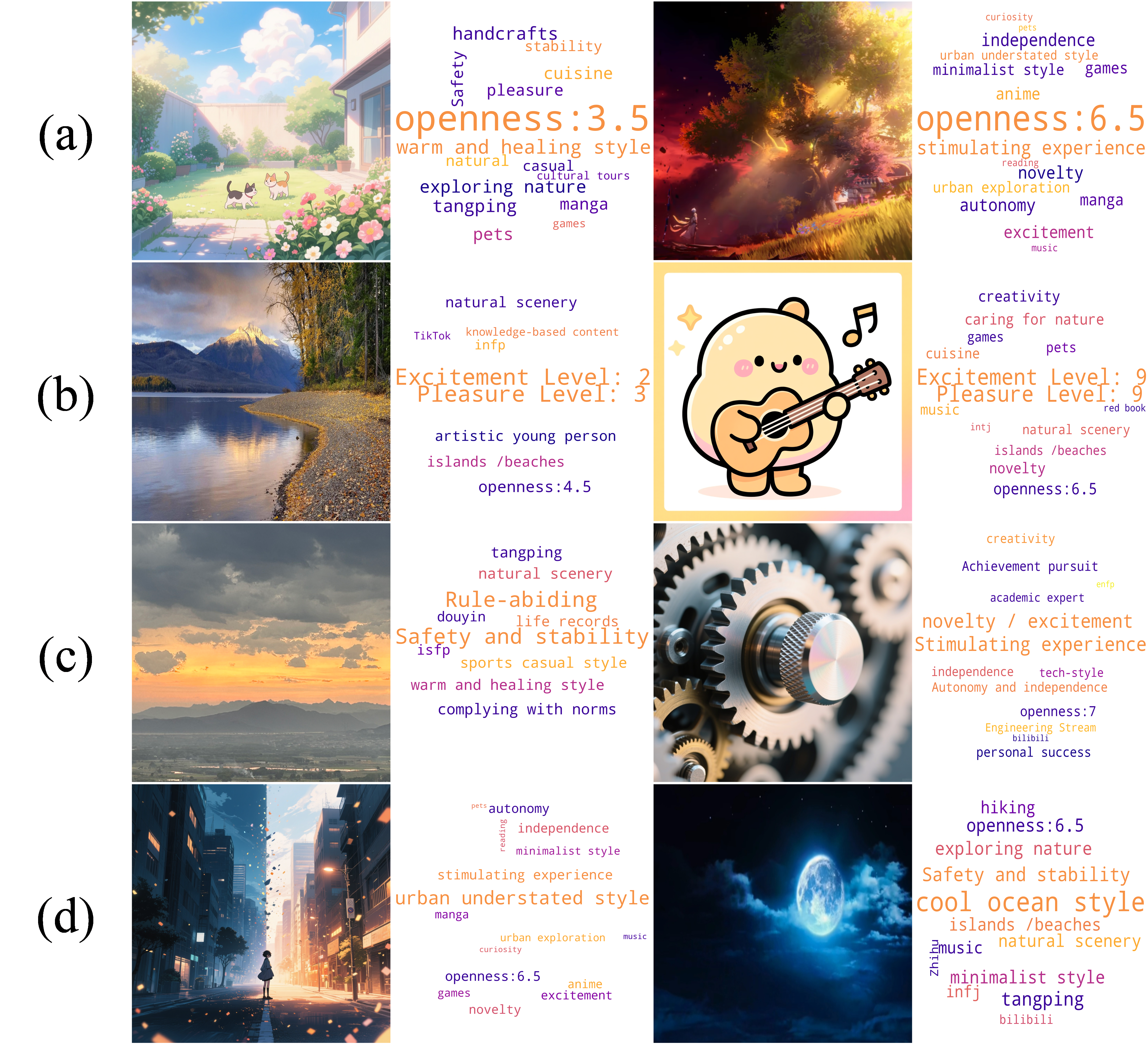}
    \caption{We used the data of User 1 and User 2 presented in Table  \ref{tab:survey}, where Column (a) shows user-uploaded images, and Columns (b) to (f) show user-selected images.}
    \label{fig:dataset3}
\end{figure}

\noindent\textbf{The second stage}.\
In the second stage, the vetted images generated for each participant were presented back to that participant in randomized order. Participants were asked to select 6–8 images they liked most as final preference annotations using the annotation interface shown in Fig.~\ref{fig:website2}. If a participant was dissatisfied with the candidate images, we regenerated images for that participant and conducted a second round of screening. In addition, participants could optionally upload extra images they personally liked as supplementary references. The user-uploaded reference images together with the finally selected preferred images are visualized in Fig.~\ref{fig:dataset3}.

In total, we collected 74 valid data entries. Each entry includes: the participant’s completed online questionnaire, the standardized user profile and image-generation prompts produced by the GPT-5 model, the set of images generated from those prompts, and the participant’s final preferred images selected from the candidate set.

\noindent\textbf{User personas example}.\
Given that the original output of GPT is redundant in expression and poorly readable, we have organized the original output of the GPT-API while ensuring that the core content remains unchanged. After organization, the profiles of the two users in Table \ref{tab:survey} are as follows:

\textbf{User1:} User 1 prefers minimalist, conceptual urban cool-style visuals. From the images she likes, it’s evident that she favors visual tension created by strong light-dark contrasts and dreamy light and shadow. Her overall style is rooted in geometric composition, negative space, and rational order—incorporating dramatic elements like light penetration, floating particles, and spatial fragmentation into restrained cool tones, thus forming a calm yet narrative visual conflict.
Her living style is "minimalist and functional," her travel preference leans toward "urban exploration/exhibition visits," she focuses on fashion and beauty in daily life, and frequently uses platforms such as Xiaohongshu (Little Red Book) and Bilibili—all indicating her high sensitivity to design sense and urban aesthetics. With an MBTI type of INTP and a thinking style of 4 (rational and neutral), she tends toward abstract thinking and structured aesthetics.
In terms of psychological measurements, her TIPI Openness score is 6.5, reflecting notable high openness, which predicts her readiness to embrace novel, experimental visual forms. Her choice of "Autonomy/Stimulation/Hedonism" in Schwartz Values further confirms her preference for innovative and exploratory images. Her CMA scores—5 for Pleasure and 3 for Excitement—reveal a preference for a restrained yet energetic visual rhythm.
Overall, her visual preferences can be summarized as: minimalism, conceptualization, urban coolness, geometric composition, sense of design, abstract/fashionable tone, along with traits of calm light and shadow, spatial contrast, and dreamy atmosphere.

\textbf{User2:} User 2 prefers realistic, warm, and healing life-like images, specifically close-ups of pets or handcrafts under natural light, quiet natural landscapes, or photos/illustrations of scenes with a historical feel. This preference stems from the fact that he has chosen a "warm and healing" living style, prefers travel focused on nature and historical culture, follows daily life-related content such as pets and games, and frequently uses platforms like Bilibili and Xiaohongshu (Little Red Book) that are dominated by lifestyle and ACG (Anime, Comics, Games) content. Additionally, his MBTI type is ISFJ, which inherently inclines him to delicate and realistic emotions, and he has a thinking style score of 4 (neutral and relatively balanced); all these traits together make him favor familiar and empathetic visual narratives.
From a psychological assessment perspective, his TIPI (Ten-Item Personality Inventory) Openness score is 3.5, indicating a tendency towards low openness and a greater preference for concrete, clear images over abstract experimental works. His selection of "Hedonism/Security and Stability" in Schwartz Values further reinforces his need for warm and secure-feeling images. Meanwhile, his current CMA (Consumer Mood Assessment) scores are 4 for Pleasure and 0 for Excitement, clearly showing his preference for a healing tone characterized by low energy, softness, and low contrast. Ultimately, his visual preferences can be condensed into core labels such as realism, healing style, life scenes, natural light, low saturation, and pet/handcraft themes.

\section{User Survey Statistics}
\label{sec:stats}

\begin{figure*}
    \centering
    \includegraphics[width=1\linewidth, keepaspectratio]{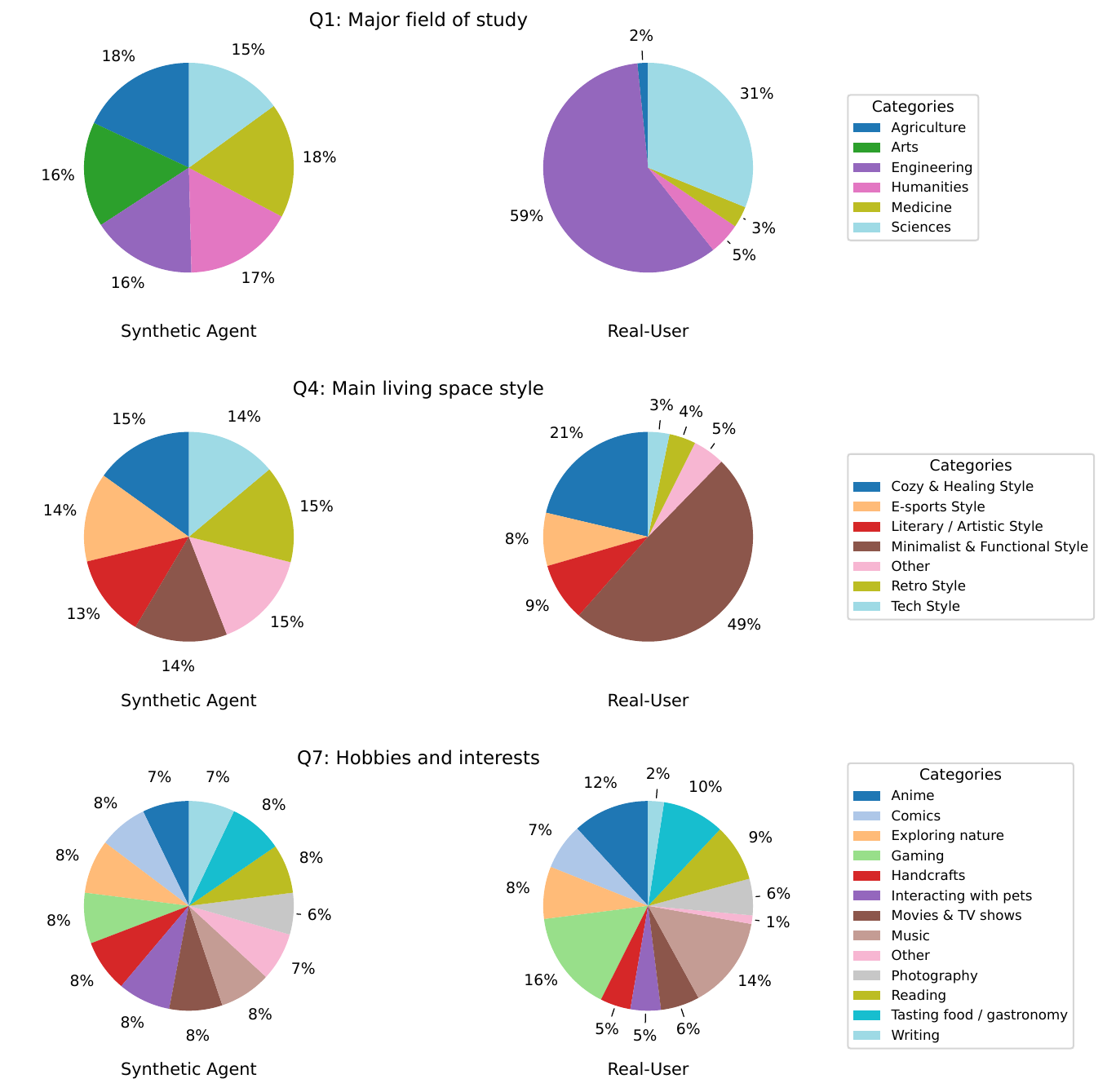}
    \caption{User survey statistics for Synthetic agents (left) and real users (right).}
    \label{fig:user_statistics}
\end{figure*}

We summarize the basic statistics of our real-user survey in Real-user dataset and synthetic agent population in Fig.~\ref{fig:user_statistics}.
For \textbf{Q1: Major field of study}, Real-user dataset respondents are clearly skewed toward STEM: the majority of participants report Engineering or Science majors, while Humanities, Agriculture, and Medical disciplines are only weakly represented. In contrast, the synthetic agents are constructed to be more balanced across fields, yielding a much flatter distribution. A similar pattern appears in \textbf{Q4: Main living space style}, where real-users strongly favor a Minimalist \& Functional style, whereas synthetic agents cover a wider mixture of cozy, literary/artistic, retro, and tech-oriented living spaces. These gaps reflect a natural bias in our recruitment pool: most of the users who are willing and able to participate in an AI–image–generation study are STEM participants with similar educational backgrounds and relatively homogeneous lifestyles.

On the other hand, \textbf{Q7: Hobbies and interests} shows a much more even spread across categories such as anime, gaming, music, reading, photography, gastronomy, and outdoor exploration. This indicates that, within the demographic we are able to recruit, we have already tried to make the survey as broad as possible and that the real-user set covers a diverse range of usage patterns and personal interests. Nevertheless, the prerequisite of “having experience with AI image-generation tools” makes it difficult to uniformly sample certain user personas, and perspectives from underrepresented groups are not fully captured. To partially mitigate this demographic imbalance, we design synthetic agents whose profiles and preferences enrich the long tail of aesthetic styles and user profiles beyond what we observe in real-user dataset. While these synthetic agents cannot fully replace collecting data from more users, they help broaden the evaluation space and reduce the impact of demographic skew. In future work, we aim to develop a more scalable, user-friendly platform that can attract a wider population of real users, thereby further improving demographic diversity and gradually reducing our reliance on synthetic augmentation.

\section{Details on Agentic Data Construction}
\label{sec:agentic}
The primary objective of this stage is to construct a larger, more evenly distributed, and psychologically coherent virtual-agent dataset from a limited number of questionnaire samples (33 in total).
Each agent must exhibit internal consistency across psychological and behavioral dimensions while maintaining sufficient diversity.
These generated agents are then converted into user-profile texts interpretable by large language models (LLMs) and, following predefined psychology–aesthetics mapping rules, used to generate semantically rich image-preference analyses and \textit{enriched prompts}.
The final outputs can be directly fed into image-generation models (\eg, Stable Diffusion) to provide human-centered, controllable inputs.
This stage consists of 3 key steps: (1) template preparation and data preprocessing, (2) agent generation, and (3) aesthetic-preference analysis and prompt generation.

\subsection{Template Preparation and Data Preprocessing}

The raw data were collected from user questionnaires consisting of 18 questions covering single-choice items (\eg, major, gender, lifestyle), multiple-choice items (\eg, interests, travel preferences, clothing styles, core values), and rating items (\eg, curiosity, rigidity, excitement).
After deduplication, outlier removal, translation, and format normalization, 33 complete and valid samples were retained and stored as the foundational data template.
This dataset defines the structure and sampling space for each question and serves as the foundational schema for subsequent agent generation.

\subsection{Agent Generation Pipeline}

The agent-generation process consists of three core modules: questionnaire parsing and encoding, random sampling and generation, and consistency/diversity validation.
The program first reads the template column headers to identify question types.
Single-choice and rating questions correspond to one column each, while multiple-choice questions allocate one column per option.
An encoder is then built for each question, converting every option into a unified ``code.label'' format (\eg, \texttt{A.Humanities}, \texttt{B.Engineering}) to maintain structural alignment with the template file.
Agents are generated through stochastic sampling: single-choice questions select one random option, multiple-choice questions draw random combinations, and rating questions sample integers within defined ranges.
Each generated agent thus forms one complete, template-aligned record.
Consistency and diversity checks are subsequently applied to ensure psychological plausibility and sample heterogeneity.

\subsection{Consistency Validation Mechanism}

Consistency validation employs a hierarchical conflict-detection and scoring system that classifies contradictions among personality traits, value orientations, and behavioral preferences into \textit{hard conflicts} and \textit{soft conflicts}.
Hard conflicts represent logically incompatible combinations and result in immediate rejection.
Examples include: high curiosity (Q12 $\ge$ 5) coexisting with high rigidity (Q13 $\ge$ 5); \textit{Self-direction} coexisting with \textit{Conformity} or \textit{Tradition}; \textit{Stimulation} coexisting with \textit{Security}; and \textit{Power} coexisting with \textit{Universalism} or \textit{Benevolence}.
Soft conflicts represent improbable but possible combinations and are penalized by a scoring system.
Each candidate starts with 100 points and loses 1 point per detected soft conflict.
These include: (A) mild mismatches between cognitive style and MBTI type, (B) weak oppositions in values (\eg, \textit{Hedonism} vs. \textit{Tradition}), (C) aesthetic–lifestyle inconsistencies (\eg, minimalism with high color saturation), and (D) AI-usage inconsistencies.
A contextual-harmonization mechanism dynamically adjusts conflict severity: \eg, when \textit{Self-direction} coexists with \textit{Conformity} and curiosity (Q12) $\ge$ 5, the conflict is downgraded; when \textit{Stimulation} coexists with \textit{Security} and excitement (Q15) $\le$ 2, it is upgraded.
An agent passes validation if it contains no hard conflicts and its cumulative score $\ge$ 97, preserving logical coherence while allowing moderate complexity and personality diversity.

\subsection{Diversity Control Mechanism}

To maintain heterogeneity among generated agents, a greedy sampling algorithm is employed with Jaccard-based diversity filtering. Each agent is represented by a \textit{signature set} that aggregates:
\begin{enumerate}
\item Multi-select answers (Q3, Q5-Q9, Q16, Q18) recorded as entries of the form QX:OptionLabel'' (\eg, \texttt{Q7:Photography}); \item Single-choice selections (\eg, field of study, gender, MBTI) encoded as Question=Option'';
\item Binned rating responses (Q11-Q15) expressed as \textit{Low}/\textit{Med}/\textit{High}.
\end{enumerate}
\noindent
When a new candidate is generated, its signature is compared against those of the most recent 200 accepted agents. The \textit{Jaccard similarity coefficient} is used to quantify overlap; if similarity with any recent signature exceeds 0.42, the candidate is rejected and regenerated. This local-window comparison maintains computational efficiency while effectively discouraging redundant answer patterns.
The final output is saved as the validated agent dataset, which satisfies three criteria: (1) logical consistency, (2) controlled diversity, and (3) template conformity.

\subsection{Aesthetic Preference Analysis and Prompt Generation}

Aesthetic analysis and prompt generation are carried out by a dedicated processing module. The template integrates psychological–aesthetic mapping rules, including TIPI openness, Schwartz values, EVT color associations, and CMA affective dimensions, allowing the LLM to jointly consider cognitive style, value orientation, and emotional state.
The model then processes the composed prompt and corresponding reference images to produce structured responses containing user metadata, inferred psychological traits, aesthetic conclusions, and up to 20 high-quality image-generation directives.

\subsection{Significance}
The two-level stage establishes a bridge between structured psychological data and computable semantic representation. By generating psychologically coherent yet diverse virtual agents, the system expands the dataset beyond real-sample limits.
Through the unified prompt framework and rule-based mapping, questionnaire responses become interpretable aesthetic descriptions and executable prompts.
This stage completes a full pipeline from \textit{questionnaire data $\rightarrow$ synthetic data $\rightarrow$ psychological/behavioral traits $\rightarrow$ aesthetic preferences $\rightarrow$ enriched prompts}, constructing a controllable, extensible, and explainable intermediate semantic layer for human-centered visual generation.

\section{Details on Image Caption Generation}
\label{sec:caption}

\begin{figure*}
    \centering
    \includegraphics[width=1\linewidth, keepaspectratio]{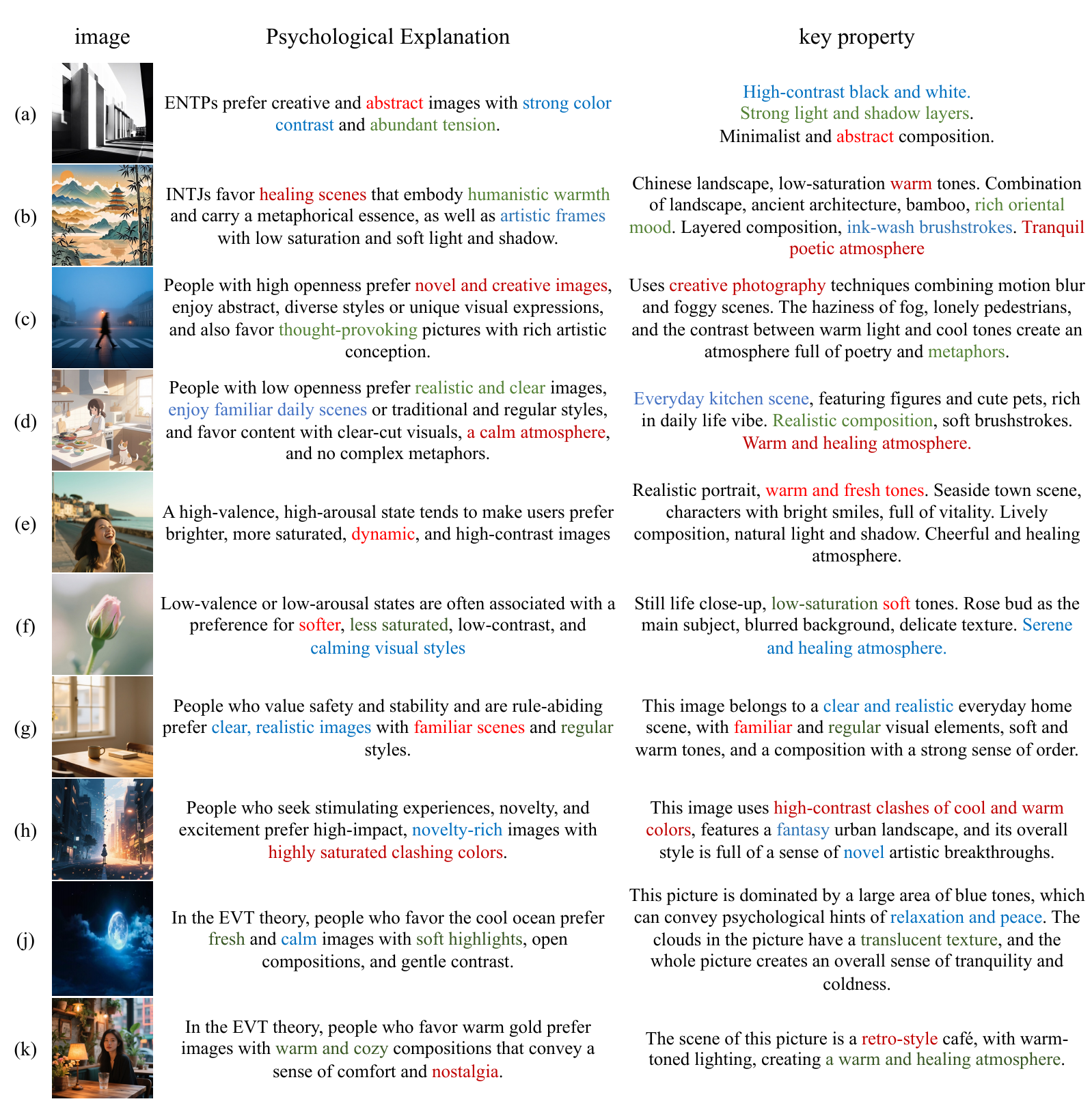}
    \caption{Examples of psychological–visual alignment in our synthetic/real-user datasets. Each row shows one of a user’s preferred image (left), a brief psychological explanation derived from their profile (middle), and the corresponding key properties (right).}
    \label{fig:partA}
\end{figure*}

\begin{figure*}
    \centering
    \includegraphics[width=1\linewidth, keepaspectratio]{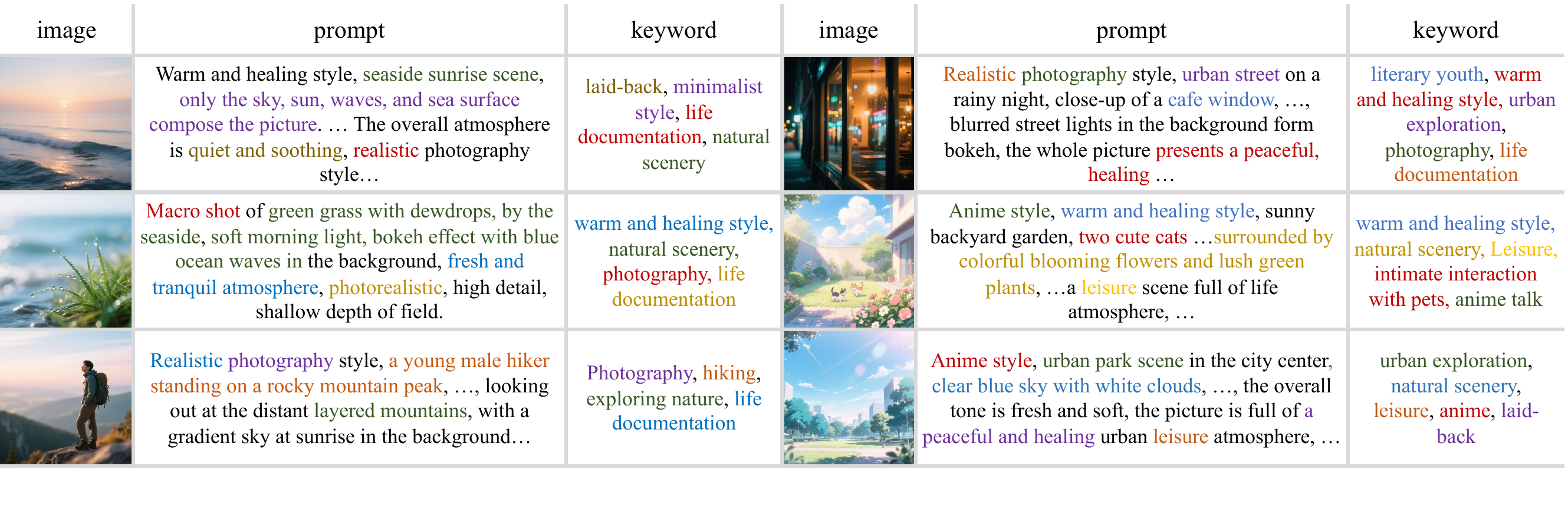}
    \caption{Examples of prompt refinement from profile keywords.
Each row starts from a set of user profile–level keywords (right). These profile cues are expanded into a refined image prompt describing concrete scene content and mood (middle), which is then used to generate the corresponding image (left). This illustrates how we translate high-level user profiles into detailed prompts and visual realizations in the synthetic/real datasets.}
    \label{fig:partB}
\end{figure*}


To ensure controllable, reproducible, and diverse reasoning, a \textbf{unified prompt framework} is implemented to connect questionnaire and image inputs to the generation of enriched image prompts. The framework consists of 6 interdependent components, each serving a distinct function within the reasoning pipeline.

\noindent\textbf{1. Instruction Header}.\ This section defines the model’s task: to infer a user’s latent image preferences by integrating questionnaire data and psychological reference tables, and to generate both a structured JSON output and 20 refined English prompts directly usable by image generation models. It delineates semantic boundaries for each question block (Q1--Q11 as explicit labels; Q12--Q15 as psychological and emotional scales) and emphasizes that the model should \textit{extract key stylistic elements} from the provided user image rather than directly copying its visual structure.

\noindent\textbf{2. Questionnaire Item List}.\ This component enumerates all referable user dimensions---discipline, gender, campus role, living style, travel and clothing preferences, hobbies, preferred content platforms, MBTI, TIPI, CMA, Schwartz values, AI usage frequency, and EVT palette choices. It also standardizes score ranges (Q11: 1--5; TIPI: 0--7; CMA: 0--9), ensuring consistent interpretation of numeric ratings during reasoning and threshold-based inference.

\noindent\textbf{3. Additional Constraint: Diversity Enhancement for Prompt Generation}.\ To prevent over-concentration of subject matter and encourage multimodal balance, the framework introduces a post-processing constraint during the generation of the 20 final prompts. It enforces:
\begin{enumerate}
\item \textbf{Balanced category distribution:} approximately 25--30\% portraits, 20\% landscapes, 15\% urban scenes, 15\% still-life, 10\% abstract visuals, and 10--15\% creative hybrids.
\item \textbf{Rotated focal subjects:} at least one prompt per major category (nature, urban, still-life, abstract, cultural/narrative).
\item \textbf{Limited repetition:} no more than two prompts may share the same dominant subject or setting.
\item \textbf{Encouraged non-human focus:} inclusion of atmospheric or object-centered scenes without explicit human subjects.
\item \textbf{Maintained stylistic coherence:} all prompts must adhere to the aesthetic profile inferred from questionnaire and image analysis (color palette, mood, realism, and compositional tone).
\end{enumerate}

\noindent\textbf{4. User Profile Slot}.\ This slot embeds the individual profile text parsed from the agent dataset, such as ``Campus Roles: \ldots / Living Space Style: \ldots / Color Preferences: \ldots.'' It serves as a bridge between structured agent data and natural language reasoning, allowing large language models to access user context.

\noindent\textbf{5. Reference Table (Theory-to-Visual Map)}.\ The reference table operationalizes psychological, behavioral, and demographic frameworks into visual reasoning rules that link human traits and contexts to aesthetic outputs. These mappings transform high-level constructs—personality, values, affect, lifestyle, and sensory preferences—into machine-controllable visual parameters for prompt generation.

\begin{itemize}
    \item \textbf{TIPI / Big Five (Openness)} $\rightarrow$ abstraction, novelty, and compositional complexity,
    \item \textbf{Schwartz’s Basic Human Values (10 values)} $\rightarrow$ narrative, thematic, and symbolic direction,
    \item \textbf{EVT (Ecological Valence Theory)} $\rightarrow$ color-affect mapping,
    \item \textbf{CMA (Circumplex Model of Affect)} $\rightarrow$ lighting, contrast, and energy,
    \item \textbf{Image-level attributes (actionable visual controls)} $\rightarrow$ prompt tokens defining generation parameters,
    \item \textbf{Age-related visual tendencies} $\rightarrow$ hue, saturation, and compositional preference,
    \item \textbf{Major and Cognitive Style} $\rightarrow$ perceptual focus and aesthetic decoding,  
    \item \textbf{Campus Role and Identity} $\rightarrow$ scenario tone and stylistic archetypes,  
    \item \textbf{Living Space Style} $\rightarrow$ compositional minimalism and material tone,  
    \item \textbf{Digital Life and Media Platforms} $\rightarrow$ cultural and subcultural visual schemas,
    \item \textbf{Hobbies and Subcultures} $\rightarrow$ symbolic iconography and thematic emphasis, 
    \item \textbf{Fashion Style Preference} $\rightarrow$ tone, texture, and subject embodiment, 
    \item \textbf{Travel Preferences} $\rightarrow$ scenic motif and spatial openness

\end{itemize}

\noindent
This extended mapping unifies psychological (TIPI, Big Five, Schwartz, EVT, CMA), demographic (age, academic background), and behavioral (media, hobbies, lifestyle) dimensions into a coherent interpretive system. It enables the LLM to translate human individuality and situational context into aesthetic variables—color, composition, lighting, realism, and style—forming the theoretical backbone of the image-preference reasoning pipeline.

\noindent\textbf{6. Output Schema (Generation Format Reference)}\. The output schema defines a reproducible data structure encompassing key psychological metrics, inferred traits (Openness, CMA, EVT, Schwartz), contextual synthesis of Q1--Q11, one-sentence style summary, actionable generation parameters (style tags, lighting, composition, constraints), scenario recommendations, and 20 executable English prompts.

\noindent\textbf{Pipeline Integration}.\ Within the overall system, these components collectively align model reasoning and enforce consistency:
\begin{itemize}
\item The \textbf{Instruction Header} defines task scope and prevents goal drift.
\item The \textbf{Questionnaire Items} establish input semantics and numeric standards.
\item The \textbf{User Profile Slot} personalizes reasoning for each agent.
\item The \textbf{Reference Table} embeds psychological--aesthetic mappings for interpretability.
\item The \textbf{Output Schema} anchors abstract reasoning into machine-usable format.
\item The \textbf{Diversity Enhancement Constraint} ensures content variety and creative balance in the generated enriched prompts.
\end{itemize}
\noindent
Collectively, this framework guarantees that every generated output is psychologically grounded, stylistically coherent, structurally reproducible, and compositionally diverse.

\section{Data Example}
\label{sec:dataexm}

\subsection{Examples of Psychological--Visual Alignment}

Fig.~\ref{fig:partA} illustrates how we connect users’ psychological profiles to their preferred images and the aesthetic key properties used in our synthetic/real data. Each row corresponds to one preferred image from an agent or real participant. The middle column gives a short psychological explanation derived from that user’s profile (\eg, Big Five traits, valence–arousal state, or EVT color theory), and the right column distills this explanation into a few concrete visual properties that we later use as interpretable aesthetic tags.

For example, case (g) shows a tidy everyday home scene. The psychological explanation describes people who value safety, stability, and rule-abiding behavior, and who therefore favor clear, realistic images with familiar scenes and regular styles. The corresponding key properties highlight exactly these cues: a clear and realistic everyday home scene with familiar visual elements and soft warm tones, reflecting order and comfort. In contrast, case (h) depicts a dynamic fantasy cityscape. Here, the profile emphasizes sensation-seeking and novelty, which is translated into key properties such as high-contrast clashes of cool and warm colors and a fantasy urban landscape with a strong sense of novel artistic breakthroughs. These examples demonstrate how psychological descriptions are systematically grounded in observable image attributes, forming the bridge between psychological traits, visual examples, and the aesthetic key properties used throughout our synthetic/real datasets.

\subsection{Examples of Prompt Refinement from Key Properties}

Fig.~\ref{fig:partB} illustrates how we translate high-level user profiles into concrete prompts and images. Each row starts from a set of profile keywords on the right. These profile cues are first expanded into a refined prompt in the middle column that specifies the details of image, and the refined prompt is then fed into the text-to-image model to generate the image on the left.

For example, in the top-left row, profile keywords such as \emph{laid-back}, \emph{minimalist style}, \emph{life documentation}, and \emph{natural scenery} are converted into a prompt describing a warm seaside sunrise where only the sky, sun, waves, and sea surface compose the picture, emphasizing a quiet and realistic photographic style. In the top-right row, keywords including \emph{literary youth}, \emph{urban exploration}, and \emph{life documentation} are refined into a prompt about a rainy-night urban street scene viewed through a cafe window, with soft lights and a peaceful, healing mood. Similarly, in the middle-right row, profile cues such as \emph{natural scenery}, \emph{anime talk}, and \emph{intimate interaction with pets} lead to an anime-style backyard garden with two cats and blooming flowers. These examples demonstrate how abstract profile descriptors are systematically grounded into detailed prompts and then into concrete visual realizations in our synthetic/real data.


\section{Additional Visualization}
\label{sec:vis}

\begin{figure*}
    \centering
    \includegraphics[width=0.9\linewidth, keepaspectratio]{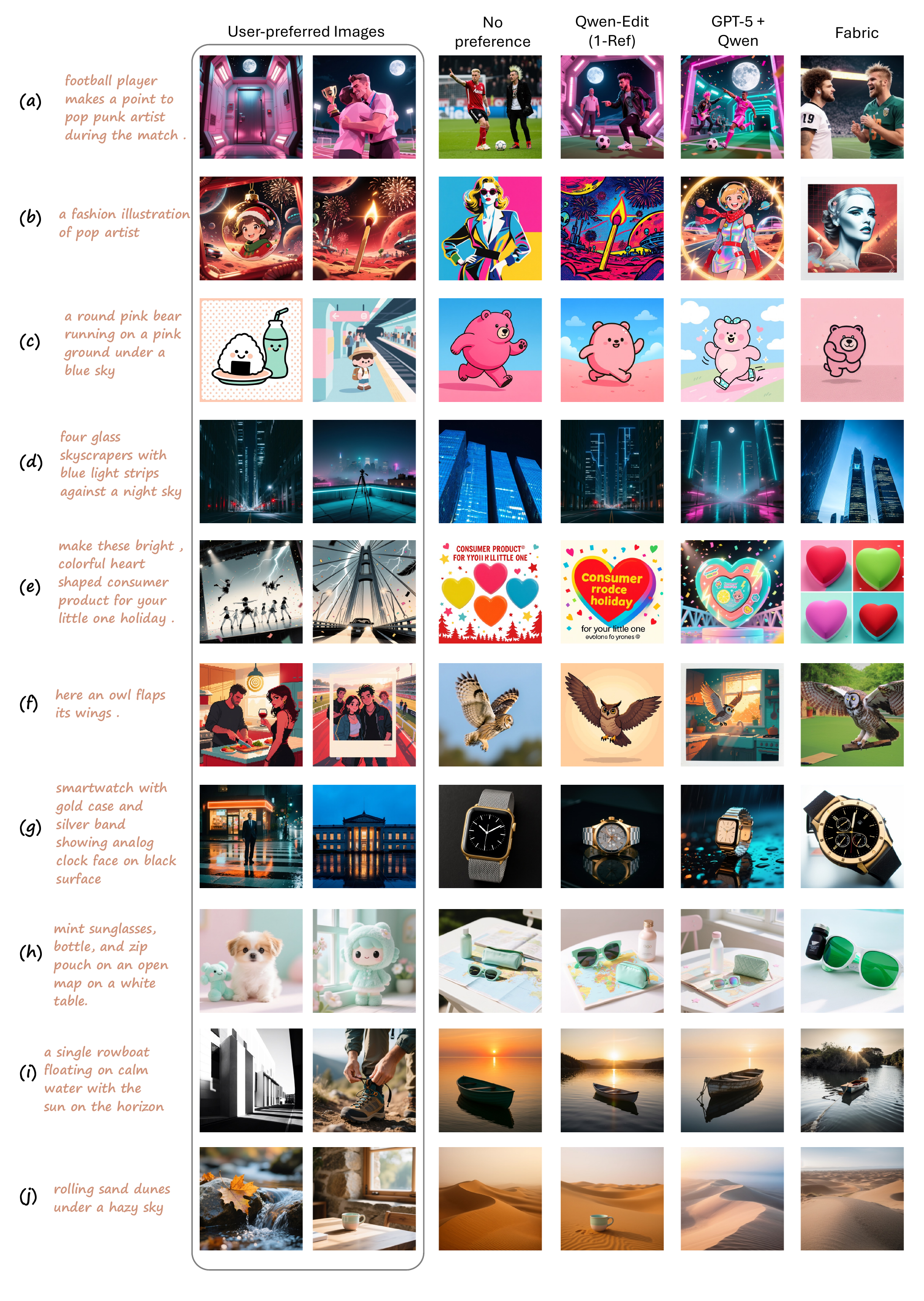}
    \caption{Additional visualizations of personalized image generation by different methods. For ease of presentation, the "User-preferred" examples display two images randomly sampled from the full reference image set.}
    \label{fig:addcomparison}
\end{figure*}

We provide additional visualizations of personalized image generation in Fig.~\ref{fig:addcomparison}, comparing the no-preference baseline, the Qwen-Image-Edit (1-Ref) joint-condition method, the GPT-5–based preference fusion method, and the Fabric separate-condition method.

\section{Reproducibility Statement}
\label{sec:reproduce}
We are committed to ensuring the reproducibility of our results. Upon acceptance and the completion of the review process, we will release (i) the PIPBench data, (ii) all code for data processing, model training, and evaluation, and (iii) configuration files and hyperparameters for all reported experiments. We will also provide the survey templates, prompt formats, and scripts used to construct the benchmark, so that future work can reproduce our pipeline and extend the dataset under similar settings.

\section{Limitations and Future Work}
\label{sec:limitations}
A natural limitation of our current dataset lies in its diversity. Most participants in our user study come from STEM backgrounds and are generally well educated, which largely aligns with the demographic of frequent image-generation users seen in real applications. However, this also means that perspectives from certain underrepresented groups are not fully reflected in the collected data. To partially address this gap, we incorporate synthetic agent profiles that extend the range of aesthetic preferences and behavioral patterns beyond those captured from real users. While synthetic agents cannot entirely replace additional user recruitment, they help broaden the evaluation space and reduce the impact of demographic imbalance on downstream analysis. In future work, we plan to develop a more scalable and user-friendly data collection platform that can attract a broader range of real users. This will allow us to further enrich demographic diversity and reduce reliance on synthetic augmentation over time.

A second limitation comes from potential dataset bias introduced by the image generator used in our pipeline. Our current system primarily relies on Qwen-Image to produce candidate images for both agent and user data collection. This is because Qwen-Image is currently the strongest open-source model available, making it a practical choice for scaling up data generation while keeping the entire pipeline reproducible. And our early studies showed that human rarely selected outputs from weaker open-source models. However, relying on a single generator may introduce mild stylistic or aesthetic bias into the candidate pool. As part of future work, we plan to expand the pipeline to incorporate candidates produced by multiple state-of-the-art proprietary models, which will help further reduce generator-specific bias and ensure a more comprehensive and diverse evaluation setting.

Beyond continued efforts to diversify and debias our benchmark, we also plan to explore more learning-based, end-to-end models that directly capture and generalize user preferences. Such models may offer stronger personalization capabilities and provide deeper insights into how preference representations can be learned from both real users and synthetic agents. We will also plan to explore how to leveraging user profiles through preference embedding or retrieval-based conditioning methods in future works.

\end{document}